\documentclass[10pt,twocolumn,letterpaper]{article}

\usepackage{cvpr}              
\definecolor{cvprblue}{rgb}{0.21,0.49,0.74}
\usepackage[pagebackref,breaklinks,colorlinks,allcolors=cvprblue]{hyperref}
\usepackage{xcolor}

\usepackage[normalem]{ulem}
\usepackage{booktabs}
\usepackage{multirow}
\usepackage{amsmath}
\usepackage{hhline}
\usepackage{csquotes}
\usepackage{tabularx}
\usepackage{booktabs}
\usepackage{makecell}

\newcolumntype{Y}{>{\centering\arraybackslash}X}

\def\paperID{} 
\def\confName{CVPR}
\def\confYear{2026}

\title{A Visual Semantic Adaptive Watermark grounded by Prefix-Tuning for Large Vision-Language Model}

\author{\textbf{Qi Zheng}\textsuperscript{\rm 1,2}$^{\ast}$,  \textbf{Shuliang Liu}\textsuperscript{\rm 1,2}$^{\ast}$, \textbf{Yu Huang}\textsuperscript{\rm 1,2}, \textbf{Sihang Jia}\textsuperscript{\rm 1,2}, \textbf{Jungang Li}\textsuperscript{\rm 1,2}, \textbf{Lyuhao Chen}\textsuperscript{\rm 3} \\  \textbf{Junhao Chen}\textsuperscript{\rm 1,2}, \textbf{Hanqian Li}\textsuperscript{\rm 1,2}, \textbf{Aiwei Liu}\textsuperscript{\rm 1,2}, \textbf{Yibo Yan}\textsuperscript{\rm 1,2}, \textbf{Xuming Hu}\textsuperscript{\rm 1,2}$^{\dagger}$\\
        \textsuperscript{\rm 1} {The Hong Kong University of Science and Technology (Guangzhou)} \\
    { \textsuperscript{\rm 2} {The Hong Kong University of Science and Technology}} \\
    { \textsuperscript{\rm 3} {Zhejiang University}}
    \\
     \texttt{\href{mailto:shulianglyo@gmail.com}{qzheng219@connect.hkust-gz.edu.cn}},
     \texttt{\href{mailto:xuminghu@hkust-gz.edu.cn}{xuminghu@hkust-gz.edu.cn}}}

\begin{document}
\maketitle
\def\thefootnote{*}\footnotetext{Equal contribution.}\def\thefootnote{\arabic{footnote}}
\def\thefootnote{†}\footnotetext{Corresponding author.}\def\thefootnote{\arabic{footnote}}
\begin{abstract}
Watermarking has emerged as a pivotal solution for content traceability and intellectual property protection in Large Vision-Language Models (LVLMs). 
However, vision-agnostic watermarks introduce visually irrelevant tokens and disrupt visual grounding by enforcing indiscriminate pseudo-random biases, while some semantic-aware methods incur prohibitive inference latency due to rejection sampling.
In this paper, we propose the \textbf{VI}sual \textbf{S}emantic \textbf{A}daptive Watermark (\textbf{VISA-Mark}), a novel framework that embeds detectable signals while strictly preserving visual fidelity. 
Our approach employs a lightweight, efficiently trained prefix-tuner to extract dynamic \textbf{Visual Evidence Weights}, which quantify the evidentiary support for candidate tokens based on the visual input.
These weights guide an adaptive vocabulary partitioning and logits perturbation mechanism, concentrating watermark strength specifically on visually-supported tokens. By actively aligning the watermark with visual evidence, VISA-Mark effectively maintains visual fidelity.
Empirical results confirm that VISA-Mark outperforms conventional methods with a 7.8\% improvement in visual consistency (Chair-I) and superior semantic fidelity. The framework maintains highly competitive detection accuracy (96.88\% AUC) and robust attack resilience (99.3\%) without sacrificing inference efficiency, effectively establishing a new standard for reliability-preserving multimodal watermarking.

\end{abstract}

\section{Introduction}
\label{sec:intro}

\begin{figure}[t!]
    \centering
    \includegraphics[width=1\linewidth]{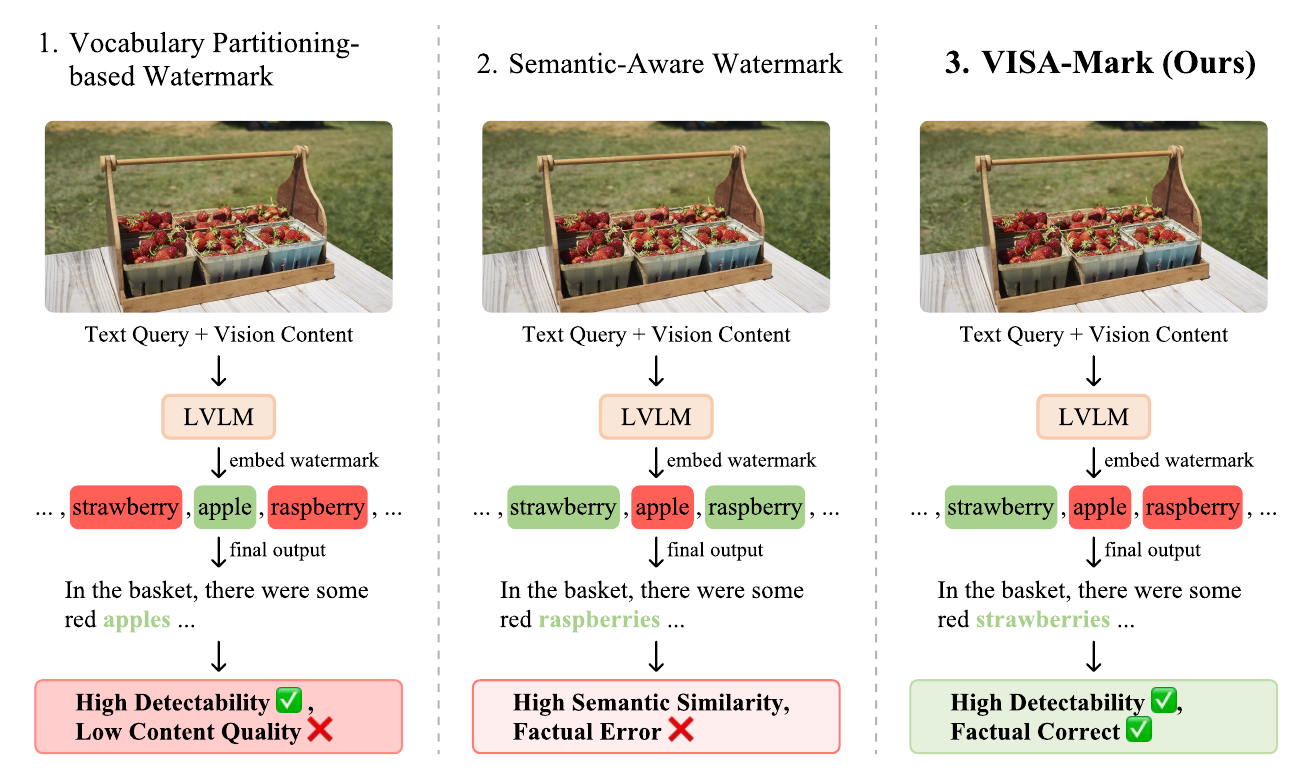}
    \caption{Paradigm comparison between our VISA-Mark and currently existing vocabulary partitioning-based watermark \& semantic-aware watermark.}
    \label{fig:introduction}
    \vspace{-0.1in}
\end{figure}

Recent breakthroughs in Large Vision-Language Models (LVLMs), such as LLaVA~\cite{liu2023visualinstructiontuning} and Qwen~\cite{bai2023qwenvlversatilevisionlanguagemodel, wang2024qwen2vlenhancingvisionlanguagemodels, bai2025qwen25vltechnicalreport}, have demonstrated remarkable capabilities in computer vision and natural language processing \cite{peng2025skywork,team2025kimi,wei2025skywork,ling2025wakenllm,zhang2025bert,zhang2025unveiling,hei2025unlocking}. 
The significant advancements in LVLMs have driven the application and transformation of technology \cite{abdelnabi2021adversarial,chen2024mark,lau2024waterfall}, but also have brought serious challenges, such as the misuse of LVLMs for malicious objectives, the proliferation of misinformation, and property right Infringement \cite{li2024unveiling,liu2024survey,nie2024securing,raz2024authorship,tang2023watermarking,chen2025safeeraser,liu2025survey}.
To solve these concerns, there is an urgent demand for a reliable method to enhance the traceability of LVLMs.
Watermarking technology \cite{li2024unveiling,liu2023unforgeable,liu2024survey,huang2025video,liu2026distilling}, which embeds imperceptible yet detectable watermarks into LLM-generated outputs \cite{dasgupta2024watermarking,gloaguen2025towards,kirchenbauer2023watermark,raz2024authorship}, has been regarded as a pivotal solution due to its potential to enhance traceability and accountability of LVLMs \cite{abdelnabi2021adversarial,antoun2023text,xu2024freqmark,zhang2024personamark,zhao2023protecting}.
The pioneering work of KGW~\cite{kirchenbauer2024watermarklargelanguagemodels} employs a pseudorandom function to partition the vocabulary and applies a positive logit bias to tokens within \enquote{green list} at each generation step \cite{chen2024mark,gloaguen2025towards,kirchenbauer2023watermark,liu2024survey,wu2024bypassing}. 
Unbiased watermarking \cite{hu2023unbiased,huo2024token,kuditipudi2023robust,mao2024watermark} maintains text quality by keeping the expected sampling distribution unchanged, but has the cost of reduced detection efficiency \cite{wu2023resilient,xie2024debiasing}.
Uncertainty-aware watermarking \cite{chen2023watme,he2024can,wang2025morphmark} enhances the robustness in low-entropy scenarios. 
Semantic-aware watermarking using contextual semantics to guide watermark injection, including textural and visual semantics \cite{liu2025vlamarkcrossmodalwatermark,huo2025pmark}. 

However, a fundamental disconnect remains, as these approaches are inherently \textbf{vision-agnostic}. They treat watermark injection as a purely linguistic probability manipulation and ignore the \textbf{visual evidence grounding}—the critical alignment ensuring generated tokens correspond to actual visual content. This oversight introduces three critical limitations when applying existing watermarking schemes to vision-language aligned generation ~\cite{nie2024securing}.
First, 
existing methodologies may create an \textbf{intrinsic conflict} between watermark injection and visual fidelity. As shown in the Fig.~\ref{fig:introduction}, vocabulary partitioning-based watermarking will break the visual consistency by introducing visually contradictory tokens, while semantic-aware watermarking confuses words with similar semantics and incorrectly increases the probability of factual error tokens \cite{tu2023waterbench,xu2025majority,liu2026vision}. 
Second, there is a contradiction between the uniform logit perturbation and detectability efficiency. Uniform logits bias spreads the same perturbation across visually grounded and irrelevant tokens, which dilutes how much bias converts into green-list probability mass, thus impairing watermark detection efficiency \cite{xu2025majority}.
Third, many semantic-aware watermarkings are based on multiple rejection sampling \cite{zhang2025cohemark,dabiriaghdam2025simmark,he2024can,mao2024watermark}, which alleviates the problems of uncertainty and consistency to some extent, but the algorithm efficiency is far lower than that of Vocabulary Partitioning-based watermarking, which limits their application in the real world \cite{min2024imitate,qu2025provably,yu2025saemark,xun2025rtv}.

\begin{figure*}[h]
    \centering
    \includegraphics[width=1\linewidth]{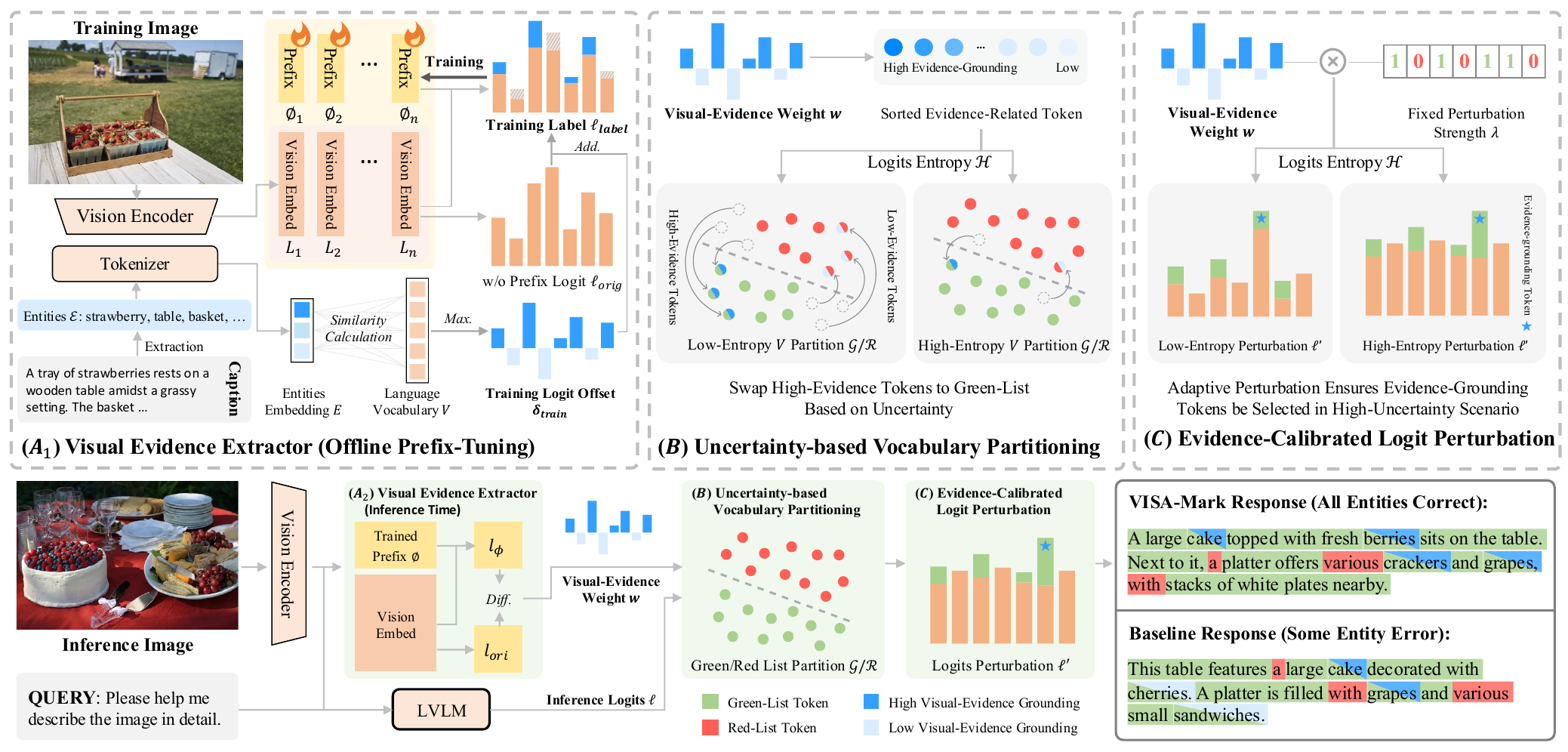}
    \caption{
    Overview of \textbf{VISA-Mark} framework, which consists of three components:
    ($A$) \textbf{Visual Evidence Extractor}: A lightweight prefix-tuner trained offline through dense image-caption pairs ($A_1$), is deployed at inference time to extract \textbf{Visual Evidence Weights} ($A_2$).
    ($B$) \textbf{Uncertainty-based Vocabulary Partitioning}: Leverages logits entropy and the extracted weights to adaptively swap high-evidence tokens into the green-list, protecting visual fidelity. 
    ($C$) \textbf{Evidence-Calibrated Logit Perturbation}: Applies a perturbation bias that scales with the Visual Evidence Weight and entropy, concentrating watermark strength on visually-grounded tokens. 
    }
    \label{fig:overview}
    \vspace{-0.1in}
\end{figure*}

To resolve these problems, we propose \textbf{VI}sual \textbf{S}emantic \textbf{A}daptive Watermark (\textbf{VISA-Mark}), a visual semantic and evidence aligned watermarking framework.
As illustrated in Fig.~\ref{fig:overview}, our approach functions through three core components: 
(\textit{A}) A \textbf{Visual Evidence Extractor}, implemented via a lightweight prefix-tuner~\cite{liu2022ptuningv2prompttuning} trained offline. This module enables the frozen LVLM to efficiently estimate dynamic visual relevance for any input image at inference, quantifying the evidentiary support for each candidate token. 
(\textit{B}) \textbf{Uncertainty-based Vocabulary Partitioning}, which safeguards visual consistency by leveraging the visual evidence weights and model uncertainty~\cite{lee2023wrote}. It preferentially swaps high-evidence tokens into the fixed-ratio green list during low-uncertainty phases, preventing the random exclusion of visually critical concepts. 
(\textit{C}) \textbf{Evidence-Calibrated Logit Perturbation}, which applies a dynamic logit bias scaled by the visual evidence weight. Instead of applying a uniform bias, this mechanism concentrates watermark strength on tokens strongly supported by the visual content.

The adaptive mechanism ensures that watermark strength is concentrated on tokens strongly supported by the vision content, actively guiding the model towards visual fidelity and away from potential hallucinations, particularly in uncertain generation steps.

Our contributions transcend prior art through three breakthroughs:
\begin{itemize}
    \item We propose a \textbf{Visual Semantic Adaptive Watermark} framework, achieving cross-modal semantic guidance through visual evidence grounding. With lightweight training overhead, it achieved a 7.8\% improvement (Chair-I $\downarrow$) in text quality and visual consistency.
    \item We developed an efficient prefix fine-tuning pipeline to extract visual evidence and implemented adaptive watermark perturbation through a visual evidence-based coordination mechanism. This two-stage visual watermarking system improves visual consistency while maintaining detection accuracy.
    \item We conducted extensive experiments to verify the effectiveness of the VISA-Mark framework in terms of text quality, visual fidelity, detectability, and robustness.
\end{itemize}

\section{Related Work}
\label{sec:related work}

\subsection{Vocabulary Partitioning-based Watermarking}

The dominant paradigm for watermarking large language models was introduced by \citet{kirchenbauer2023watermark}, which pseudorandomly partitions the vocabulary into a \enquote{green list} at each step and applies a fixed logit bias to embed a detectable signal. Many subsequent works have built upon this foundation, aiming to improve text quality, statistical properties, or robustness. These include methods for unbiased or distribution-preserving watermarking \citep{wu2023resilient, hu2023unbiased, xie2024debiasing, mao2024watermark}, strategies to enhance multi-bit capacity or robustness against attacks \citep{xu2025majority, qu2025provably, wang2023towards, lau2024waterfall}, and alternative partitioning schemes based on neural networks or sinusoidal signals \citep{zhao2023protecting, liu2023unforgeable}.

A fundamental limitation, as noted in surveys \citep{liu2024survey} and analyses \citep{rastogi2024revisiting}, is that these approaches are inherently \textbf{content-agnostic}, or more critically for multimodal tasks, \textbf{vision-agnostic}. By indiscriminately applying a bias, they risk suppressing visually-grounded tokens that fall outside the random green list, which can, as our work shows, exacerbate model hallucinations. Even methods designed for other data types, like tabular data \citep{he2024watermarking}, rely on statistical partitioning rather than semantic consistency.

\subsection{Semantic-Aware and Context-Guided Watermarking}

To address the quality degradation of random partitioning, another line of work has explored semantic-aware watermarking. However, the vast majority of these methods are designed for unimodal text. They leverage textual cohesion \citep{zhang2025cohemark}, lexical redundancy (synonyms) \citep{chen2023watme}, textual context embeddings \citep{liu2310semantic, hu2022hiure}, cross-lingual semantics \citep{he2024can}, or linguistic features like keywords and syntax \citep{yoo2023robust}. While improving textual fidelity, these approaches remain vision-agnostic and fail to align the watermark with visual evidence.

Other methods adapt the watermark based on the model's predictive uncertainty (entropy) \citep{wang2025morphmark, lee2023wrote, lu2024entropy, zhang2025catmark}, but do not consider the \textbf{visual relevance} of tokens. A different category employs post-hoc rejection sampling or rewriting \citep{yu2025saemark, dabiriaghdam2025simmark, chang2024postmark, li2025treehop}, which can introduce significant inference latency and cannot guide the initial generation toward visual fidelity. Techniques designed for code \citep{li2024acw}, end-to-end rewriting \citep{abdelnabi2021adversarial, zhang2024remark}, or embedding models \citep{tang2023watermarking} are not directly applicable to guiding the token-by-token generative process of LVLMs to maintain visual-semantic alignment.

\subsubsection{Prefix-Tuning} 
Prefix-tuning~\cite{liu2021pretrainpromptpredictsystematic} represents an important paradigm in Parameter-Efficient Fine-Tuning (PEFT), enabling the adaptation of Large Pre-trained Models (PLMs) by optimizing a small, continuous prefix vector while keeping backbone parameters frozen~\cite{li2021prefixtuningoptimizingcontinuousprompts}.
This methodology has demonstrated efficacy comparable to full fine-tuning across diverse natural language processing tasks~\cite{liu2022ptuningv2prompttuning, tam2022parameterefficientprompttuningmakes, yang2022robustprefixtuningtextclassification, lester2021powerscaleparameterefficientprompt, liu2023gptunderstands, wang2025prefixtuningmodernizingprefixtuningdecoupling, ouyang-etal-2023-prefix} and multimodal applications~\cite{jia2022visualprompttuning, liu2022fewshotparameterefficientfinetuningbetter, 10657279, yao2023visuallanguageprompttuningknowledgeguided}. Despite its success, prior research has predominantly utilized prefix-tuning for downstream task adaptation. Its potential as a modular, inference-time mechanism to steer internal generative processes, specifically for extracting dynamic evidence weights, remains largely underexplored.


Our work, VISA-Mark, is the first to bridge this critical gap. 
It introduces a watermarking framework that is not only vision-aware but also \textbf{vision-adaptive}, using prefix-tuning as a \textbf{visual evidence extractor} to dynamically guide the watermark embedding process. This allows it to simultaneously ensure robust detectability and actively maintain visual fidelity, resolving the core conflict between reliability and traceability in LVLMs.

\section{Methodology}
\label{sec:methodology}

%

We propose \textbf{VISA-Mark}, a vision-aligned watermarking framework that estimates token-level \emph{Visual Evidence Weights} (VEW) to align watermark injection with visual-grounded semantics.
Our method is built from three components (Fig.~\ref{fig:overview}): 
(i) a prefix-tuned extractor that produces dense, bounded VEW without modifying backbone weights (Sec.~\ref{sec:visual_evidence}); 
(ii) an uncertainty-regulated vocabulary partition that swaps high-evidence tokens into the green list while keeping overall green list size fixed for detection (Sec.~\ref{sec:entropy_adaptive}); and 
(iii) an evidence-calibrated logit perturbation that scales bias by VEW to ensure that token selection is aligned with visual evidence 
(Sec.~\ref{sec:adaptive_perturbation}).
Together, these modules preserve the detector’s null statistics
, yielding strong detectability with improved visual fidelity compared to vision-agnostic schemes.

\subsection{Problem Setup}
\label{sec:preliminary}




Let $\mathcal{M}$ be a frozen Large Vision–Language Model (LVLM) with vocabulary $\mathcal{V}$ of size $|\mathcal{V}|$.
Given a visual input $\mathbf{v}$ and a text prefix $y_{1:t-1}$, the next-token distribution is
\begin{equation}
    p_t = \mathrm{softmax}\!\big(\ell_t\big), \qquad \ell_t \triangleq \mathcal{M}(y_{1:t-1},\mathbf{v}) \in \mathbb{R}^{\mathcal{V}}, 
\label{eq:llminference}
\end{equation}
Classical red/green (R/G) watermarks perturb logits with a hash key $s_t$:
$\hat{p}_t=\mathcal{F}(p_t,s_t)$.
We hypothesize that a \emph{vision-agnostic} perturbation $\mathcal{F}$ may conflict with the visual grounding learned by $\mathcal{M}$, harming visual consistency and text quality, which is consistent with our experimental results in Sec.~\ref{sec:main_result}.
We therefore introduce \emph{visual evidence weights} $\omega(i)\in(0,1)$ for each token $i\in\mathcal{V}$ and design a \emph{vision-aware} perturbation
\begin{equation}
    \hat{p}_t=\mathcal{F}'\!\big(p_t, s_t, \omega\big),
\end{equation}
which (i) aligns injected bias with visual evidence and (ii) adapts to model uncertainty.

\subsection{Component \textit{A}: Visual Evidence Weight Extracting}
\label{sec:visual_evidence}

Our first challenge is to acquire the visual evidence weights $\boldsymbol{\omega}$ efficiently. Methods like full fine-tuning are computationally prohibitive and undesirably modify frozen model parameters, while external neural networks lack portability and are difficult to align with the LVLM's internal knowledge. To avoid these issues, we adopt a more parameter-efficient approach following P-Tuning~\cite{liu2022ptuningv2prompttuning}. 
We first train a prefix using an offline pipeline, which then serves as a modular extractor during the inference phase.
In the offline prefix-tuning phase, as shown in Fig.~\ref{fig:overview} ($A_1$), we capture fine-grained relationships between visual content and linguistic vocabulary using external knowledge.  
Then train a small, lightweight dummy prefix $\phi$ to guide the frozen LVLM in generating the desired visual evidence weights $\boldsymbol{\omega} \in \mathbb{R}^\mathcal{V}$. This prefix is used to extract visual evidence as an external component in the pre-watermarking process, demonstrated in Fig.~\ref{fig:overview} ($A_2$).

\subsubsection{Offline Prefix-Tuning Pipeline}

We leverage a dense image--caption corpus $\{(x_m,c_m)\}_{m=1}^{M}$ from DCI dataset~\cite{urbanek2024pictureworth77text} as external knowledge, where $c_m$ denotes the $m$th caption. For each image--caption pair, we summarize the visual evidence as a set of entities $\mathcal{E}_m=\{e_{m,k}\}_{k=1}^{K_m}$ extracted from the caption by Part-of-Speech tags:
\begin{equation}
    \mathcal{E}_m \;=\; \big\{\, \chi \in \mathrm{Chunks}(c_m) \;:\; \mathrm{ChunkTag}(\chi)=\mathrm{NP} \,\big\},
\end{equation}
where $\mathrm{Chunks}(c_m)$ is the set of phrase chunks and $\mathrm{NP}$ denotes noun phrases. Each entity is then embedded as $E_{m,k}=\text{Tokenizer}(e_{m,k})$.

To capture visually relevant lexical variants beyond this limited entity set, we compute a visual-linguistic relevance score $s_i$ for each token $i \in \mathcal{V}$ by comparing it with the entity embeddings:
\begin{equation}
    s_i=\max_{k} \sigma(E_{m,k},\mathbf{u}_i), \  \text{where} \  \sigma(E,\mathbf{u})=\frac{E^{\top}\mathbf{u}}{\lVert E\rVert \cdot \lVert \mathbf{u}\rVert},
\end{equation}
where $\{\mathbf{u}_i\}_{i=1}^{\mathcal{V}}$ are the embedded language vocabulary. For each token $i$, $s_i$ is the maximum cosine similarity to any embedding entity. This process produces a dense weight vector $\mathbf{s} \in \mathbb{R}^\mathcal{V}$ that reflects the visual relevance of the entire language vocabulary based on the input image.
We convert relevance scores $\mathbf{s}$ into logit offsets $\boldsymbol{\delta}_{\text{train}}$  for training:
\begin{equation}
\delta_i \;=\; \mathrm{clip}\!\left( \tilde s_i,\,-1,\,1\right),
\quad
\boldsymbol{\delta}_{\text{train}}=[\delta_1,\ldots,\delta_{|\mathcal{V}|}]^{\top},
\label{eq:delta_clip}
\end{equation}
where $\tilde s_i$ is normalized by $\tilde s_i=(s_i-\mu_s)/\sigma_s$.
Let $\ell_{\text{orig}} = \mathcal{M}(\mathbf{v})$ be the base model's single-step inference logits with only vision input. We form the \emph{target label logits} $ \ell_{\text{label}}$ by adding our computed offset:
\begin{equation}
    \ell_{\text{label}} \;=\; \ell_{\text{orig}} \;+\; \kappa \cdot \boldsymbol{\delta}_{\text{train}},
\label{eq:train_label}
\end{equation}
where $\kappa$ controls the strength of the logit offset in the training process. 

We attach the virtual prefix $\phi$ and obtain prefix-conditioned logits $\ell^{(\phi)} = \mathcal{M}(\mathbf{v}, \phi)$. The prefix is trained to match the \emph{target distribution} $L$ via a temperatured KL divergence objective:
\begin{equation}
\begin{aligned}
\mathcal{L}
= \sum_{t}
\mathrm{KL}\!\Big(
\mathrm{softmax}( \ell_{\text{label}}/\tau)
&\,\big\|\,
\mathrm{softmax}(\ell^{(\phi)}/\tau)
\Big)
\\
&\quad + \lambda_{\text{reg}} \,\lVert \phi\rVert_2^{2},
\end{aligned}
\label{eq:kl_loss}
\end{equation} 

where $\tau$ is a temperature and $\lambda_{\text{reg}}$ controls prefix regularization. During training, gradients flow only through $\phi$; all base model parameters remain frozen.  

As shown in Fig.~\ref{fig:overview} ($A_1$), this pipeline consolidates discrete visual entities extracted from captions into a dense, vocabulary-wide distribution for prefix tuning. We effectively distill visual-linguistic correlations into a lightweight module without the computational overhead of full fine-tuning. Crucially, this transforms the otherwise sparse and implicit supervisory signals of raw text into a comprehensive global prior, ensuring the model captures a broader spectrum of visually relevant concepts.

\subsubsection{Inference Phase Extractor}

During inference time, we deploy the trained prefix $\phi$ as an efficient visual evidence extractor module.
We employ a contrastive decoding strategy \cite{wang2024mitigating} to extract the dynamic visual-token weights. Given the input vision content $\mathbf{v}$, we compute two logit vectors in parallel:
\begin{itemize}
    \item $\ell_{\text{orig}} \in \mathbb{R}^{\mathcal{V}}$: original logits from $\mathcal{M}(\mathbf{v})$ (without $\phi$).
    \item $\ell^{(\phi)} \in \mathbb{R}^{\mathcal{V}}$: prefix-conditioned logits from $\mathcal{M}(\phi, \mathbf{v})$.
\end{itemize}
We define the contrastive logit difference $\Delta \ell(i) = \ell^{(\phi)}(i) - \ell_{\text{orig}}(i)$. This difference $\Delta \ell(i)$ quantifies the influence of the prefix vector: a high positive value indicates that token $i$ emphasizes visual evidence alignment. We normalize these differences to serve as our bounded weights $w(i) \in (0,1)$:
\begin{equation}
    w(i) = \operatorname{sigmoid}\!\left(\frac{\Delta \ell(i) - \mu}{\sigma}\right),
\label{eq:weights_normalized}
\end{equation}
where $\mu$ and $\sigma$ denote the mean and standard deviation of the logit differences, respectively.

It is worth noting that this module operates with \textbf{constant computational overhead}. Since the weights are derived solely from the static visual input, they are computed only once at the initial stage. As a result, the inference cost remains invariant to the number of generated tokens, guaranteeing that the pipeline maintains high efficiency even for long-text generation. A detailed quantitative analysis of inference latency is provided in Appendix C.

\subsection{Component \textit{B}: Uncertainty-based Vocabulary Partitioning}
\label{sec:entropy_adaptive}

The model infers the probability value $p_t$ of the next token based on the given visual and text input, as shown in Eq.~\ref{eq:llminference}. To enhance text quality and visual consistency while maintaining watermark detectability, we utilize token entropy as an uncertainty metric to adaptively adjust the vocabulary partitioning mechanism.

At each time step $t$, we measure the token entropy $H_t$:
\begin{equation}
    \mathcal{H}_t = -\sum_{i=1}^{\mathcal{V}} p_{t, i}\,\log p_{t, i}, 
\end{equation}

The normalized entropy, which quantifies the uncertainty at each generation step, is then determined by:

\begin{equation}
    \mathcal{H}_{\text{norm}} = \frac{\mathcal{H}_{t}}{\mathcal{H}_{\text{max}}} = \frac{\mathcal{H}_{t}}{\log |\mathcal{V}|},
\label{eq:normalized_entropy}
\end{equation}
where $\mathcal{H}_{\text{max}}$ is the theoretical maximum value of entropy \cite{liu2025vlamarkcrossmodalwatermark}.
Based on the normalized entropy $\mathcal{H}_{\text{norm}}$, we calculate the evidence-grounding tokens ratio $\eta_t$:
\begin{equation}
    \eta_{t} = \alpha(1 - \mathcal{H}_{\text{norm}}),
\end{equation}
where the \textbf{Evidence-Grounded Token Ratio} $\alpha$ controls the base evidence-grounding token proportion. We keep the ratio of green-list fixed as $\gamma=0.5$ as \citet{kirchenbauer2023watermark}. Let $\mathcal G_t$ (green) and $\mathcal R_t$ (red) be the PRF-seeded partition at step $t$. We form a candidate set $\mathcal{C}_t$, which selects the tokens with the highest visual evidence weights:
\begin{equation}
    \mathcal{C}_t = \underset{i \in \mathcal{V}}{\operatorname{arg\,TopK}}\left( w(i), \, \left\lceil \eta_t \mathcal{V} \right\rceil \right),
\end{equation}
where $\mathcal{C}_t$ consists of the top $\left\lceil \eta_t \mathcal{V} \right\rceil$ tokens (a proportion $\eta_t$ of the total vocabulary $\mathcal{V}$) selected from the vocabulary $\mathcal{V}$ based on the highest standardized visual weights $w(i)$.
We then swap $A_t=\mathcal C_t\cap\mathcal R_t$ into green by removing the $|A_t|$ least-evidence tokens $B_t\subset\mathcal G_t$:
\begin{equation}
\begin{split}
    \mathcal G_t\leftarrow(\mathcal G_t\setminus B_t)\cup A_t, \\
    \mathcal R_t\leftarrow(\mathcal R_t\setminus A_t)\cup B_t,
\end{split}
\end{equation}
optionally gating the swap by a margin threshold and a per-step cap to avoid oscillation.

This adaptive partitioning resolves the conflict between detectability and visual consistency by preventing the random red list $\mathcal R_t$ from penalizing visually-grounded tokens. 
The uncertainty-aware ratio $\eta_t$ dynamically regulates this process: expanding visual evidence inclusion during low-entropy steps to maximize fidelity, while prioritizing stochastic partitioning in high-entropy steps for robustness. Crucially, by maintaining an invariant green list size, our method enhances visual alignment without compromising the statistical integrity of the detector’s null distribution.


\subsection{Component \textit{C}: Evidence-Calibrated Logits Perturbation}
\label{sec:adaptive_perturbation}
A standard watermark applies a uniform bias, which can be suboptimal. This may lead to the selection of visually irrelevant tokens, compromising visual consistency. To address this, we reformulate the logit perturbation to be \textbf{evidence-calibrated} and \textbf{uncertainty-aware}.

To achieve evidence-calibrated perturbation, for each token $v \in \mathcal{G}_t$ in our dynamic green list, we first introduce a token-specific regulating factor $\psi_{t,v}$, 
which dynamically scales the perturbation intensity by incorporating both model uncertainty, from the normalized entropy $\mathcal{H}_{\text{norm}}$ from Eq.~\ref{eq:normalized_entropy}, and visual grounding, from the visual relevance weight, respectively:

\begin{equation}
    \psi_{t, v} = \beta \cdot \mathcal{H}_{\text{norm}}\cdot w(v),
\end{equation}
where $\beta$ is a hyperparameter controlling the global \textbf{logits perturbation strength}.

We compute the final positive logits bias $\delta_{t,v}$, which is formulated by modulating the fixed base bias $\lambda = 0.5$ with the regulating factor $\psi_{t,v}$:
\begin{equation}
    \delta_{t,v} = \lambda \cdot \psi_{t, v}  + \lambda, \quad \forall v \in \mathcal{G}_t,
    \label{eq:adaptive_delta}
\end{equation}
where $\lambda$ is the fixed bias.
This formulation ensures that the watermark signal always maintains a baseline intensity of $\lambda$, while receiving an adaptive boost $\lambda \cdot \psi_{t,v}$ that is proportional to both the generation uncertainty and the token's visual evidence. Finally, the perturbed logits $\ell'_{t}$ are obtained by applying this adaptive bias $\delta_{t,v}$ exclusively to the green list $\mathcal{G}_t$, while applying neutral treatment to the red list $\mathcal{R}_t$.
\begin{equation}
    \ell'_{t,v} = \begin{cases} 
      \ell_t(v) + \delta_{t,v} & \text{if } v \in \mathcal{G}_t, \\
      \ell_t(v) & \text{if } v \notin \mathcal{G}_t.
   \end{cases}
   \label{eq:final_logits}
\end{equation}

This evidence-calibrated mechanism achieves a dual purpose. 
First, by scaling the perturbation $\delta_{t,v}$ with the visual evidence weight $w(v)$, we concentrate watermark strength on visually grounded tokens while minimizing disturbances to weakly relevant ones, thereby preserving visual fidelity. 
Second, the entropy regulation $\mathcal{H}_{\text{norm}}$ dynamically adapts the bias intensity: it applies stronger, evidence-aligned perturbations during high-uncertainty steps to suppress hallucinations, while relaxing the bias during low-uncertainty phases to maintain robust detectability.

\begin{table*}[t]
\centering
\footnotesize 
\caption{Performance comparison of \textbf{VISA-M} against baseline watermarking methods on the LLaVA and Qwen models, evaluated on the MS-COCO 14, MS-COCO 17, and AMBER benchmarks. Metrics include watermark detectability (AUC $\uparrow$), text quality (PPL $\downarrow$ and BertScore $\uparrow$), and visual consistency (Chair-I $\downarrow$). VISA-M consistently achieves superior visual consistency and text quality while maintaining highly competitive detection accuracy. \textbf{Bold} values indicate the best performance among all methods, while \underline{underlined} indicate the second best. 
`NW' denotes the 'No Watermark' baseline and is excluded from best/second-best highlighting.
}
\begin{tabular}{llcccccccccccc} 
\toprule
 & & \multicolumn{4}{c}{\textbf{MS-COCO 14}} & \multicolumn{4}{c}{\textbf{MS-COCO 17}} & \multicolumn{4}{c}{\textbf{AMBER}} \\
\cmidrule(lr){3-6}\cmidrule(lr){7-10}\cmidrule(lr){11-14}
Model & Method & AUC & PPL & BertScore & Chair-I & AUC & PPL & BertScore & Chair-I & AUC & PPL & BertScore & Chair-I \\ 
\midrule
\multirow{9}{*}{\textbf{LLaVA}} 
 & NW & / & 5.24 & / & 16.26 & / & 5.23 & / & 16.81 & / & 5.60 & / & 18.09 \\
\cmidrule(lr){2-14}
 & VLA & 89.29 & 5.80 & 92.79 & 17.94 & 88.22 & 5.81 & 92.58 & \underline{16.68} & 88.54 & 6.04 & 92.80 & 18.80 \\
 & KGW & 95.70 & 5.83 & 92.70 & 17.37 & 95.57 & 5.79 & 92.66 & 16.98 & 95.39 & \underline{6.08} & 92.74 & \underline{18.03} \\
 & SWEET & \underline{96.50} & \underline{5.74} & 92.69 & 19.25 & \underline{96.10} & \underline{5.69} & 92.65 & 20.05 & \textbf{95.80} & 6.10 & 92.73 & 30.15 \\
 & DiP & 84.37 & 5.92 & 92.91 & \underline{16.91} & 74.44 & 5.70 & 93.72 & 17.53 & 87.48 & 6.38 & 92.82 & 18.61 \\
 & Unbiased & 84.33 & 5.94 & \underline{92.96} & 17.57 & 74.41 & 5.70 & \underline{93.78} & 17.15 & 86.41 & 6.36 & \underline{92.86} & 18.75 \\
\cmidrule(lr){2-14}
 & \textbf{VISA-M} & \textbf{97.95} & \textbf{5.52} & \textbf{93.07} & \textbf{16.39} & \textbf{98.05} & \textbf{5.59} & \textbf{93.80} & \textbf{16.15} & \underline{95.51} & \textbf{5.91} & \textbf{92.92} & \textbf{17.25} \\
\midrule
\multirow{9}{*}{\textbf{Qwen}} 
 & NW & / & 3.01 & / & 6.65 & / & 3.01 & / & 7.34 & / & 2.98 & / & 11.30 \\
\cmidrule(lr){2-14}
 & VLA & 82.18 & \underline{3.05} & 93.49 & 6.53 & 78.44 & \underline{3.08} & 93.57 & 7.48 & 78.45 & \textbf{3.03} & 93.80 & 12.15 \\
 & KGW & \underline{82.44} & 3.08 & 93.62 & 6.18 & \underline{80.71} & 3.11 & 93.67 & \underline{7.12} & \underline{81.35} & 3.06 & 93.85 & 11.96 \\
 & SWEET & 76.76 & 3.15 & \underline{94.14} & \underline{6.02} & 72.27 & 3.18 & \underline{93.90} & 7.31 & 76.42 & 3.15 & \underline{94.50} & 12.60 \\
 & DiP & 77.78 & 3.15 & 93.25 & 6.27 & 74.09 & 3.13 & 93.82 & 7.55 & 78.85 & 3.13 & 93.44 & \underline{11.63} \\
 & Unbiased & 77.67 & 3.16 & 93.24 & 6.21 & 73.60 & 3.14 & 93.52 & 7.19 & 78.86 & 3.13 & 93.45 & 11.71 \\
\cmidrule(lr){2-14}
 & \textbf{VISA-M} & \textbf{84.53} & \textbf{3.04} & \textbf{94.67} & \textbf{5.68} & \textbf{84.21} & \textbf{3.02} & \textbf{94.31} & \textbf{7.10} & \textbf{82.97} & \textbf{3.03} & \textbf{94.60} & \textbf{11.42} \\
\bottomrule
\end{tabular}
\label{tab:performance_metrics}
\end{table*}

\section{Experiment}
\label{sec:experiment}

Our experiments comprehensively assessed VISA-Mark’s performance against five baseline methods on AMBER~\cite{wang2024amberllmfreemultidimensionalbenchmark}, MS-COCO 14 and 17~\cite{lin2015microsoftcococommonobjects} datasets, focusing on three primary areas: (1) text quality and visual fidelity, (2) watermark detectability, and (3) robustness. We conducted an ablation study to evaluate the individual contributions of our core components: the Uncertainty-based Vocabulary Partitioning component and the Evidence-Calibrated Logits Perturbation component. Additionally, we assessed VISA-Mark's resilience against a suite of textual attacks to confirm its robustness.

\subsection{Experiment Setup}

\paragraph{Models and datasets.}
Our approach is assessed on two state-of-the-art large vision-language models: LLaVA-v1.5~\cite{liu2023visualinstructiontuning} and Qwen3-VL~\cite{bai2025qwen25vltechnicalreport, wang2024qwen2vlenhancingvisionlanguagemodels, bai2023qwenvlversatilevisionlanguagemodel}. 
Additionally, we trained the respective prefix vectors for these two vision-language models using our prefix training pipeline, with detailed results provided in Appendix A.

\paragraph{Baselines.}
 Our approach compares with five representative watermark baselines: KGW~\cite{kirchenbauer2024watermarklargelanguagemodels}, SWEET~\cite{lee2024wrotecodewatermarkingcode}, Unbiased~\cite{hu2023unbiasedwatermarklargelanguage}, DiP~\cite{wu2024resilientaccessibledistributionpreservingwatermark} and VLA-Mark~\cite{liu2025vlamarkcrossmodalwatermark} using MarkLLM~\cite{pan2024markllmopensourcetoolkitllm} repository with the official hyperparameter.
 In fairness, we fix the same sampling policy and length budget between methods.

\paragraph{Evaluation Metrics.}
Our evaluation spans detectability performance (AUC and Accuracy), 
visual consistency (Chair-I), 
text quality (PPL and BertScore), 
and robustness against three types of attack, which are altering text through word insertion, deletion, and synonym substitution.

\begin{figure*}[h]
    \centering
    \includegraphics[width=1\linewidth]{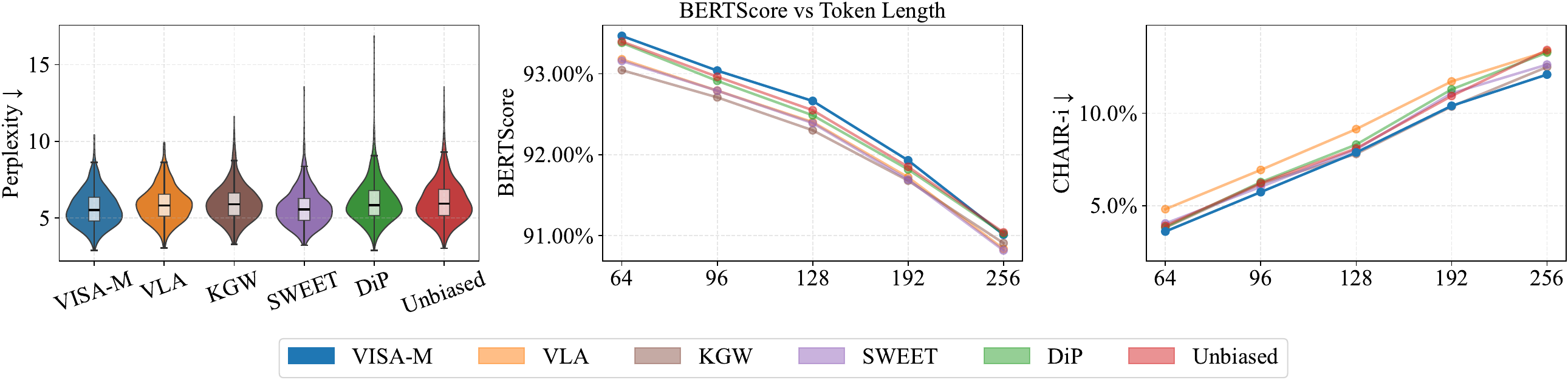}
    \caption{Text quality and visual consistency analysis between VISA-Mark and baseline methods. \textbf{Left:} Violin plots of perplexity scores; VISA-M shows a lower median and tighter distribution, indicating higher fluency. \textbf{Middle:} BERTScore versus token length; our method mitigates semantic degradation in long-text generation. \textbf{Right:} Chair-I versus token length; VISA-M maintains the lowest hallucination rate as generation grows, confirming robust visual fidelity.}
    \label{fig:text_quality}
    \vspace{-0.1in}
\end{figure*}

\subsection{Main Results}

\subsubsection{Watermark}
\label{sec:main_result}

Table~\ref{tab:performance_metrics} presents a comprehensive quantitative comparison between VISA-Mark and five baseline methods across LLaVA and Qwen models. The results empirically validate our primary hypothesis: while vision-agnostic watermarking mechanisms degrade visual consistency and text quality, our vision-adaptive approach actively preserves and enhances them. Additional case studies are provided in Appendix D.

As illustrated in Table~\ref{tab:performance_metrics}, VISA-Mark demonstrates a superior balance across the critical tripartite trade-off of detection accuracy, text quality, and visual consistency. 
Specifically, our method achieves consistent best performance in text quality metrics (PPL and BertScore) and visual fidelity (Chair-I) across all configurations. For instance, on the LLaVA backbone, VISA-Mark reduces the Chair-I score on MS-COCO 14 to \textbf{16.39}, significantly outperforming the standard watermark KGW (17.37) and semantic-aware watermark VLA (17.94). 
Crucially, these improvements do not come at the cost of security. VISA-Mark maintains high detection accuracy, achieving the highest AUC on almost all experience settings. 
This confirms that embedding visual evidence into the watermarking process effectively aligns the generated text with visual content without compromising the watermark's statistical detectability.

This balanced performance stems from our dual mechanism of visual evidence alignment and entropy regulation. 
By dynamically modulating watermark strength according to model confidence, VISA-Mark ensures robust detectability during high-confidence (low-entropy) phases while preventing the inadvertent exclusion of visually grounded tokens. 
Conversely, in high-uncertainty states where visual consistency is fragile, the mechanism explicitly prioritizes the selection of visually aligned tokens. This strategy effectively mitigates hallucination risks while preserving the semantic integrity of the generated text.

\begin{figure*}[h]
    \centering
    \includegraphics[width=1\linewidth]{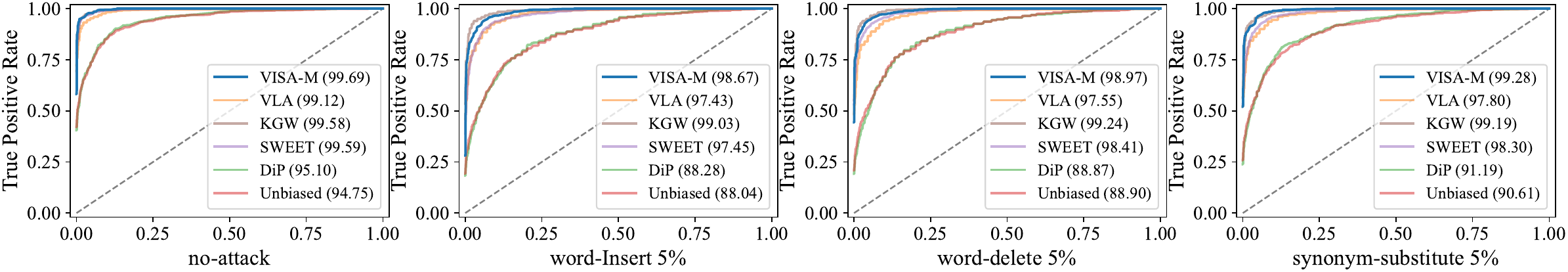}
    \caption{ROC curves evaluating detection performance under no-attack and three text attack scenarios (Word-Insert, Word-Delete, Synonym-Substitute at 5\% rate). VISA-Mark (blue curve) demonstrates superior robustness, maintaining near-perfect AUC across all attacks, whereas baselines like DiP and Unbiased exhibit performance collapse.}
    \label{fig:robustness}
    \vspace{-0.1in}
\end{figure*}

\subsubsection{Ablation Study}
\begin{table}[t]
\centering
\footnotesize 
\caption{Ablation study on the evidence-grounded token ratio $\alpha$ and perturbation strength $\beta$. The results demonstrate a trade-off between detectability and visual fidelity. The configuration $\alpha=0.005, \beta=0.5$ achieves the optimal balance, minimizing hallucinations (Chair-I) and improving text quality (PPL and BertScore) while maintaining high detection AUC and Accuracy.}
\begin{tabular*}{\linewidth}{@{\extracolsep{\fill}}lcccccc}
\toprule
 Ablation of $\alpha$ & 0.0 & 0.0025 & 0.005* & 0.0075 & 0.01 & 0.02 \\ 
\midrule
 AUC & \textbf{96.91} & 94.93 & 93.99 & 92.74 & 92.20 & 89.54  \\
 Accuracy & \textbf{90.65} & 87.45 & 85.8 & 84.95 & 83.90 & 82.30  \\
 PPL $\downarrow$ & 5.86 & 5.69 & 5.70 & 5.69 & 5.68 & \textbf{5.61}  \\
 BertScore & 92.62 & 92.85 & 92.84 & 92.88 & 92.91 & \textbf{93.04}  \\
 Chair-I $\downarrow$ & 16.79 & \textbf{15.49} & 15.76 & 15.80 & 15.74 & 15.62  \\
\midrule
\midrule
 Ablation of $\beta$ & 0.0 & 0.25 & 0.5* & 0.75 & 1.0 & 2.0 \\ 
\midrule
 AUC & \textbf{95.24} & 95.14 & 94.38 & 93.71 & 93.19 & 90.56  \\
 Accuracy & \textbf{88.30} & 87.85 & 87.06 & 86.55 & 85.30 & 82.05  \\
 PPL $\downarrow$ & 5.74 & 5.72 & \textbf{5.69} & 5.71 & \textbf{5.69} & 5.78  \\
 BertScore & \textbf{92.88} & 92.86 & 92.85 & \textbf{92.88} & 92.80 & 92.57  \\
 Chair-I $\downarrow$ & 16.74 & 16.27 & 15.52 & 15.43 & \textbf{14.61} & 15.43  \\
\bottomrule
\end{tabular*}
\label{tab:ablation}
\end{table}

We investigate the impact of two critical hyperparameters: the evidence-grounded token ratio $\alpha$ and the logits perturbation strength $\beta$, which regulate the \textit{Uncertainty-based Vocabulary Partitioning} and \textit{Evidence-Calibrated Logits Perturbation} components, respectively. Additional ablation studies are presented in Appendix B.

As illustrated in Table~\ref{tab:ablation}, both hyperparameters exhibit a distinct trade-off between detectability (AUC and Accuracy) and generation fidelity (PPL, BertScore, and Chair-I). Specifically, increasing $\alpha$ and $\beta$ consistently improves text quality and reduces hallucinations (\textit{e.g.}, PPL and Chair-I drop to 5.69 and 14.61 when $\beta=1.0$). This validates our component design: a higher $\alpha$ allows more visually grounded tokens to bypass the random red-list exclusion, while a larger $\beta$ increases the probability of these evidence-rich tokens, effectively enforcing visual consistency.

However, the results also highlight that excessive values for either parameter compromise detection performance. Over-prioritizing semantic tokens or applying aggressive perturbations disrupts the statistical randomness required for the watermark detector, leading to a decline in detectability efficiency (\textit{e.g.}, AUC and Accuracy drops to 89.54\% and 82.30\% when $\alpha=0.02$). Consequently, we identify the configuration of $\alpha=0.005$ and $\beta=0.5$ as the optimal equilibrium. This setting maintains robust detectability ($\text{AUC}\approx94-96\%$) while achieving minimal perplexity and optimal visual alignment, demonstrating the robustness of our method to hyperparameter selection.



\subsubsection{Text Quality Maintenance and Visual Semantic Fidelity}

We further analyze the impact of watermarking on text quality and visual fidelity across varying generation lengths.

In Figure~\ref{fig:text_quality} (Left), the violin plots reveal that VISA-Mark exhibits a lower median perplexity with a more concentrated distribution compared to baselines like KGW and DiP. This indicates that our watermarked text remains closer to the natural language distribution of the original model. This advantage stems from our \textit{Visual Evidence Weighting} mechanism, which protects visually correct tokens from being arbitrarily rejected by the random partitioning process, ensuring that perturbations are only applied where they do not disrupt linguistic fluency.

As shown in Figure~\ref{fig:text_quality} (Middle), while semantic similarity (BERTScore) naturally degrades across all methods as the generation length increases from 64 to 256 tokens, VISA-Mark consistently maintains superior performance. This suggests that our dynamic token exchange strategy from the \textit{Uncertainty-based Vocabulary Partitioning} component effectively minimizes the "semantic drift" often observed in long-context watermarking. By prioritizing tokens crucial to the overall visual narrative, we preserve the global coherence of the generated description.

Crucially, Figure~\ref{fig:text_quality} (Right) highlights the impact on visual consistency. As the sequence length grows, the cumulative probability of hallucination (Chair-I) rises for all models. However, VISA-Mark consistently achieves the lowest rate of hallucination. This demonstrates that our \textit{Evidence-Calibrated Logit Perturbation} effectively anchors the generation to the visual input. By providing stronger reinforcement to evidence-aligned tokens, our method prevents the "hallucination snowballing" effect, ensuring high fidelity even in longer responses.



\subsubsection{Detectability and Robustness}

To assess the resilience of our watermark, we evaluated VISA-Mark against three standard text-space attacks: random word insertion, deletion, and synonym substitution. Following standard protocols, we modified 5\% of the tokens in generated responses.

Figure~\ref{fig:robustness} presents the ROC curves and AUC metrics. In the pristine "no-attack" scenario, VISA-Mark achieves state-of-the-art detectability with an AUC of \textbf{99.69\%}, surpassing all competitive baselines.
Crucially, VISA-Mark exhibits exceptional robustness when subjected to adversarial attacks. While baselines such as DiP and Unbiased suffer a significant performance collapse, experiencing an average AUC drop of approximately \textbf{7\%} into the 88\%--91\% range, VISA-Mark maintains robust detectability with minimal degradation. Specifically, the AUC retains \textbf{98.97\%} performance level under Insertion, \textbf{99.27\%} under Deletion, and \textbf{99.59\%} under Synonym Substitution. VISA-Mark achieves this exceptional robustness without sacrificing visual consistency.

We attribute this resilience to our core visual evidence anchoring strategy.
By prioritizing visually grounded tokens, VISA-Mark ensures the watermark signal remains invariant across meaning-preserving attacks. Specifically, since synonyms share high relevance, they consistently receive probability boosts, preserving the signal during substitution. 
Furthermore, by anchoring the watermark to deterministic and content-critical concepts, VISA-Mark maintains signal integrity against structural attacks such as insertion and deletion, establishing a highly robust paradigm for multimodal watermarking.

\section{Conclusion and Limitation}
\label{sec:limitation}

We have presented \textbf{VISA-Mark}, a visual semantic adaptive watermarking framework that harmonizes content authenticity with cross-modal information fidelity. By synergizing a prefix-based visual-evidence extractor, uncertainty-regulated vocabulary partitioning, and evidence-calibrated logit perturbation, our method balances detection efficiency and visual semantic consistency. Empirical results demonstrate VISA-Mark's superiority, achieving competitive detectability and high robustness while improving visual fidelity and text quality. This work establishes a vision-adaptive paradigm, ensuring that watermark injection reinforces rather than disrupts visual grounding.

Despite these advancements, limitations remain. 
First, the prefix-tuner's reliance on dense caption training data may influence generalization to highly out-of-distribution domains, such as medical imaging or abstract art, particularly in the absence of domain-specific adaptation.
Second, while VISA-Mark exhibits strong resistance to common text-space attacks, its vulnerability to adaptive attacks specifically targeted the evidence-extraction mechanism warrants further study. 
Finally, our current pipeline, which extracts evidence primarily from noun phrases, focuses on object-level evidence; extending the framework to mitigate fine-grained attribute or relational inconsistencies remains a critical direction for future work.

{
    \small
    \bibliographystyle{ieeenat_fullname}
    \bibliography{main}

@String(CVPR= {IEEE Conf. Comput. Vis. Pattern Recog.})

@String(CVPR  = {CVPR})

@misc{liu2023visualinstructiontuning,
      title={Visual Instruction Tuning}, 
      author={Haotian Liu and Chunyuan Li and Qingyang Wu and Yong Jae Lee},
      year={2023},
      eprint={2304.08485},
      archivePrefix={arXiv},
      primaryClass={cs.CV},
      url={https://arxiv.org/abs/2304.08485}, 
}

@misc{bai2023qwenvlversatilevisionlanguagemodel,
      title={Qwen-VL: A Versatile Vision-Language Model for Understanding, Localization, Text Reading, and Beyond}, 
      author={Jinze Bai and Shuai Bai and Shusheng Yang and Shijie Wang and Sinan Tan and Peng Wang and Junyang Lin and Chang Zhou and Jingren Zhou},
      year={2023},
      eprint={2308.12966},
      archivePrefix={arXiv},
      primaryClass={cs.CV},
      url={https://arxiv.org/abs/2308.12966}, 
}

@misc{wang2024qwen2vlenhancingvisionlanguagemodels,
      title={Qwen2-VL: Enhancing Vision-Language Model's Perception of the World at Any Resolution}, 
      author={Peng Wang and Shuai Bai and Sinan Tan and Shijie Wang and Zhihao Fan and Jinze Bai and Keqin Chen and Xuejing Liu and Jialin Wang and Wenbin Ge and Yang Fan and Kai Dang and Mengfei Du and Xuancheng Ren and Rui Men and Dayiheng Liu and Chang Zhou and Jingren Zhou and Junyang Lin},
      year={2024},
      eprint={2409.12191},
      archivePrefix={arXiv},
      primaryClass={cs.CV},
      url={https://arxiv.org/abs/2409.12191}, 
}

@misc{bai2025qwen25vltechnicalreport,
      title={Qwen2.5-VL Technical Report}, 
      author={Shuai Bai and Keqin Chen and Xuejing Liu and Jialin Wang and Wenbin Ge and Sibo Song and Kai Dang and Peng Wang and Shijie Wang and Jun Tang and Humen Zhong and Yuanzhi Zhu and Mingkun Yang and Zhaohai Li and Jianqiang Wan and Pengfei Wang and Wei Ding and Zheren Fu and Yiheng Xu and Jiabo Ye and Xi Zhang and Tianbao Xie and Zesen Cheng and Hang Zhang and Zhibo Yang and Haiyang Xu and Junyang Lin},
      year={2025},
      eprint={2502.13923},
      archivePrefix={arXiv},
      primaryClass={cs.CV},
      url={https://arxiv.org/abs/2502.13923}, 
}

@misc{kirchenbauer2024watermarklargelanguagemodels,
      title={A Watermark for Large Language Models}, 
      author={John Kirchenbauer and Jonas Geiping and Yuxin Wen and Jonathan Katz and Ian Miers and Tom Goldstein},
      year={2024},
      eprint={2301.10226},
      archivePrefix={arXiv},
      primaryClass={cs.LG},
      url={https://arxiv.org/abs/2301.10226}, 
}

@misc{liu2025vlamarkcrossmodalwatermark,
      title={VLA-Mark: A cross modal watermark for large vision-language alignment model}, 
      author={Shuliang Liu and Qi Zheng and Jesse Jiaxi Xu and Yibo Yan and Junyan Zhang and He Geng and Aiwei Liu and Peijie Jiang and Jia Liu and Yik-Cheung Tam and Xuming Hu},
      year={2025},
      eprint={2507.14067},
      archivePrefix={arXiv},
      primaryClass={cs.CV},
      url={https://arxiv.org/abs/2507.14067}, 
}

@misc{wu2024resilientaccessibledistributionpreservingwatermark,
      title={A Resilient and Accessible Distribution-Preserving Watermark for Large Language Models}, 
      author={Yihan Wu and Zhengmian Hu and Junfeng Guo and Hongyang Zhang and Heng Huang},
      year={2024},
      eprint={2310.07710},
      archivePrefix={arXiv},
      primaryClass={cs.CR},
      url={https://arxiv.org/abs/2310.07710}, 
}

@misc{lee2024wrotecodewatermarkingcode,
      title={Who Wrote this Code? Watermarking for Code Generation}, 
      author={Taehyun Lee and Seokhee Hong and Jaewoo Ahn and Ilgee Hong and Hwaran Lee and Sangdoo Yun and Jamin Shin and Gunhee Kim},
      year={2024},
      eprint={2305.15060},
      archivePrefix={arXiv},
      primaryClass={cs.CL},
      url={https://arxiv.org/abs/2305.15060}, 
}

@inproceedings{xun2025rtv,
title={{RTV}-Bench: Benchmarking {MLLM} Continuous Perception, Understanding and Reasoning through Real-Time Video},
author={ShuHang Xun and Sicheng Tao and Jungang Li and Yibo Shi and Zhixin Lin and Zhanhui Zhu and Yibo Yan and Hanqian Li and LingHao Zhang and Shikang Wang and Yixin Liu and Hanbo Zhang and Ying Ma and Xuming Hu},
booktitle={The Thirty-ninth Annual Conference on Neural Information Processing Systems Datasets and Benchmarks Track},
year={2025},
url={https://openreview.net/forum?id=MpaSsvHcWC}
}

@misc{hu2023unbiasedwatermarklargelanguage,
      title={Unbiased Watermark for Large Language Models}, 
      author={Zhengmian Hu and Lichang Chen and Xidong Wu and Yihan Wu and Hongyang Zhang and Heng Huang},
      year={2023},
      eprint={2310.10669},
      archivePrefix={arXiv},
      primaryClass={cs.CR},
      url={https://arxiv.org/abs/2310.10669}, 
}

@misc{urbanek2024pictureworth77text,
      title={A Picture is Worth More Than 77 Text Tokens: Evaluating CLIP-Style Models on Dense Captions}, 
      author={Jack Urbanek and Florian Bordes and Pietro Astolfi and Mary Williamson and Vasu Sharma and Adriana Romero-Soriano},
      year={2024},
      eprint={2312.08578},
      archivePrefix={arXiv},
      primaryClass={cs.CV},
      url={https://arxiv.org/abs/2312.08578}, 
}

@misc{pan2024markllmopensourcetoolkitllm,
      title={MarkLLM: An Open-Source Toolkit for LLM Watermarking}, 
      author={Leyi Pan and Aiwei Liu and Zhiwei He and Zitian Gao and Xuandong Zhao and Yijian Lu and Binglin Zhou and Shuliang Liu and Xuming Hu and Lijie Wen and Irwin King and Philip S. Yu},
      year={2024},
      eprint={2405.10051},
      archivePrefix={arXiv},
      primaryClass={cs.CR},
      url={https://arxiv.org/abs/2405.10051}, 
}

@misc{lin2015microsoftcococommonobjects,
      title={Microsoft COCO: Common Objects in Context}, 
      author={Tsung-Yi Lin and Michael Maire and Serge Belongie and Lubomir Bourdev and Ross Girshick and James Hays and Pietro Perona and Deva Ramanan and C. Lawrence Zitnick and Piotr Dollár},
      year={2015},
      eprint={1405.0312},
      archivePrefix={arXiv},
      primaryClass={cs.CV},
      url={https://arxiv.org/abs/1405.0312}, 
}

@misc{wang2024amberllmfreemultidimensionalbenchmark,
      title={AMBER: An LLM-free Multi-dimensional Benchmark for MLLMs Hallucination Evaluation}, 
      author={Junyang Wang and Yuhang Wang and Guohai Xu and Jing Zhang and Yukai Gu and Haitao Jia and Jiaqi Wang and Haiyang Xu and Ming Yan and Ji Zhang and Jitao Sang},
      year={2024},
      eprint={2311.07397},
      archivePrefix={arXiv},
      primaryClass={cs.CL},
      url={https://arxiv.org/abs/2311.07397}, 
}

@misc{liu2022ptuningv2prompttuning,
      title={P-Tuning v2: Prompt Tuning Can Be Comparable to Fine-tuning Universally Across Scales and Tasks}, 
      author={Xiao Liu and Kaixuan Ji and Yicheng Fu and Weng Lam Tam and Zhengxiao Du and Zhilin Yang and Jie Tang},
      year={2022},
      eprint={2110.07602},
      archivePrefix={arXiv},
      primaryClass={cs.CL},
      url={https://arxiv.org/abs/2110.07602}, 
}

@inproceedings{abdelnabi2021adversarial,
 author = {Abdelnabi, Sahar and Fritz, Mario},
 booktitle = {2021 IEEE Symposium on Security and Privacy (SP)},
 organization = {IEEE},
 pages = {121--140},
 title = {Adversarial watermarking transformer: Towards tracing text provenance with data hiding},
 venue = {2021 IEEE Symposium on Security and …},
 year = {2021}
}

@article{antoun2023text,
 author = {Antoun, Wissam and Sagot, Beno{\^\i}t and Seddah, Djam{\'e}},
 journal = {arXiv preprint arXiv:2309.13322},
 title = {From text to source: Results in detecting large language model-generated content},
 venue = {arXiv preprint arXiv:2309.13322},
 year = {2023}
}

@article{chang2024postmark,
 author = {Chang, Yapei and Krishna, Kalpesh and Houmansadr, Amir and Wieting, John and Iyyer, Mohit},
 journal = {arXiv preprint arXiv:2406.14517},
 title = {Postmark: A robust blackbox watermark for large language models},
 venue = {arXiv preprint arXiv …},
 year = {2024}
}

@article{chen2023watme,
 author = {Chen, Liang and Bian, Yatao and Deng, Yang and Cai, Deng and Li, Shuaiyi and Zhao, Peilin and Wong, Kam-Fai},
 journal = {arXiv preprint arXiv:2311.09832},
 title = {WatME: Towards lossless watermarking through lexical redundancy},
 venue = {arXiv preprint arXiv …},
 year = {2023}
}

@article{chen2024mark,
 author = {Chen, Ruibo and Wu, Yihan and Guo, Junfeng and Huang, Heng},
 journal = {arXiv preprint arXiv:2410.13808},
 title = {De-mark: Watermark Removal in Large Language Models},
 venue = {arXiv preprint arXiv:2410.13808},
 year = {2024}
}

@article{dabiriaghdam2025simmark,
 author = {Dabiriaghdam, Amirhossein and Wang, Lele},
 journal = {arXiv preprint arXiv:2502.02787},
 title = {Simmark: A robust sentence-level similarity-based watermarking algorithm for large language models},
 venue = {arXiv preprint arXiv:2502.02787},
 year = {2025}
}

@article{dasgupta2024watermarking,
 author = {Dasgupta, Agnibh and Tanvir, Abdullah and Zhong, Xin},
 journal = {arXiv preprint arXiv:2411.05091},
 title = {Watermarking language models through language models},
 venue = {arXiv preprint arXiv:2411.05091},
 year = {2024}
}

@article{gloaguen2025towards,
 author = {Gloaguen, Thibaud and Jovanovi{\'c}, Nikola and Staab, Robin and Vechev, Martin},
 journal = {arXiv preprint arXiv:2502.10525},
 title = {Towards Watermarking of Open-Source LLMs},
 venue = {arXiv preprint arXiv …},
 year = {2025}
}

@article{he2024can,
 author = {He, Zhiwei and Zhou, Binglin and Hao, Hongkun and Liu, Aiwei and Wang, Xing and Tu, Zhaopeng and Zhang, Zhuosheng and Wang, Rui},
 journal = {arXiv preprint arXiv:2402.14007},
 title = {Can watermarks survive translation? on the cross-lingual consistency of text watermark for large language models},
 venue = {arXiv preprint arXiv …},
 year = {2024}
}

@article{he2024watermarking,
 author = {He, Hengzhi and Yu, Peiyu and Ren, Junpeng and Wu, Ying Nian and Cheng, Guang},
 journal = {arXiv preprint arXiv:2405.14018},
 title = {Watermarking generative tabular data},
 venue = {arXiv preprint arXiv:2405.14018},
 year = {2024}
}

@article{hu2023unbiased,
 author = {Hu, Zhengmian and Chen, Lichang and Wu, Xidong and Wu, Yihan and Zhang, Hongyang and Huang, Heng},
 journal = {arXiv preprint arXiv:2310.10669},
 title = {Unbiased watermark for large language models},
 venue = {arXiv preprint arXiv …},
 year = {2023}
}

@article{huo2024token,
 author = {Huo, Mingjia and Somayajula, Sai Ashish and Liang, Youwei and Zhang, Ruisi and Koushanfar, Farinaz and Xie, Pengtao},
 journal = {arXiv preprint arXiv:2402.18059},
 title = {Token-specific watermarking with enhanced detectability and semantic coherence for large language models},
 venue = {arXiv preprint arXiv …},
 year = {2024}
}

@inproceedings{kirchenbauer2023watermark,
 author = {Kirchenbauer, John and Geiping, Jonas and Wen, Yuxin and Katz, Jonathan and Miers, Ian and Goldstein, Tom},
 booktitle = {International Conference on Machine Learning},
 organization = {PMLR},
 pages = {17061--17084},
 title = {A watermark for large language models},
 venue = {International …},
 year = {2023}
}

@article{kuditipudi2023robust,
 author = {Kuditipudi, Rohith and Thickstun, John and Hashimoto, Tatsunori and Liang, Percy},
 journal = {arXiv preprint arXiv:2307.15593},
 title = {Robust distortion-free watermarks for language models},
 venue = {arXiv preprint arXiv …},
 year = {2023}
}

@article{lau2024waterfall,
 author = {Lau, Gregory Kang Ruey and Niu, Xinyuan and Dao, Hieu and Chen, Jiangwei and Foo, Chuan-Sheng and Low, Bryan Kian Hsiang},
 journal = {arXiv preprint arXiv:2407.04411},
 title = {Waterfall: Framework for Robust and Scalable Text Watermarking and Provenance for LLMs},
 venue = {arXiv preprint arXiv …},
 year = {2024}
}

@article{lee2023wrote,
 author = {Lee, Taehyun and Hong, Seokhee and Ahn, Jaewoo and Hong, Ilgee and Lee, Hwaran and Yun, Sangdoo and Shin, Jamin and Kim, Gunhee},
 journal = {arXiv preprint arXiv:2305.15060},
 title = {Who wrote this code? watermarking for code generation},
 venue = {arXiv preprint arXiv …},
 year = {2023}
}

@article{li2024acw,
 author = {Li, Boquan and Zhang, Mengdi and Zhang, Peixin and Sun, Jun and Wang, Xingmei and Fu, Zirui},
 journal = {arXiv preprint arXiv:2402.07518},
 title = {ACW: Enhancing Traceability of AI-Generated Codes Based on Watermarking},
 venue = {arXiv preprint arXiv …},
 year = {2024}
}

@article{li2024unveiling,
 author = {Li, Zhiying and Liu, Zhi and Liu, Dongjie and Zhuo, Shengda and Geng, Guanggang and Weng, Jian and Lyu, Shanxiang and Jin, Xiaobo},
 journal = {arXiv preprint arXiv:2411.15450},
 title = {Unveiling the Achilles' Heel: Backdoor Watermarking Forgery Attack in Public Dataset Protection},
 venue = {arXiv preprint arXiv …},
 year = {2024}
}

@article{liu2023unforgeable,
 author = {Liu, Aiwei and Pan, Leyi and Hu, Xuming and Li, Shu'ang and Wen, Lijie and King, Irwin and Yu, Philip S},
 journal = {arXiv preprint arXiv:2307.16230},
 title = {An unforgeable publicly verifiable watermark for large language models},
 venue = {arXiv preprint arXiv …},
 year = {2023}
}

@article{liu2024survey,
 author = {Liu, Aiwei and Pan, Leyi and Lu, Yijian and Li, Jingjing and Hu, Xuming and Zhang, Xi and Wen, Lijie and King, Irwin and Xiong, Hui and Yu, Philip},
 journal = {ACM Computing Surveys},
 number = {2},
 pages = {1--36},
 publisher = {ACM New York, NY},
 title = {A survey of text watermarking in the era of large language models},
 venue = {ACM Computing …},
 volume = {57},
 year = {2024}
}

@article{liu2310semantic,
 author = {Liu, Aiwei and Pan, Leyi and Hu, Xuming and Meng, Shiao and Wen, Lijie},
 journal = {URL https://arxiv. org/abs/2310.06356},
 title = {A semantic invariant robust watermark for large language models, 2024},
 venue = {NA},
 year = {NA}
}

@article{lu2024entropy,
 author = {Lu, Yijian and Liu, Aiwei and Yu, Dianzhi and Li, Jingjing and King, Irwin},
 journal = {arXiv preprint arXiv:2403.13485},
 title = {An entropy-based text watermarking detection method},
 venue = {arXiv preprint arXiv:2403.13485},
 year = {2024}
}

@article{mao2024watermark,
 author = {Mao, Minjia and Wei, Dongjun and Chen, Zeyu and Fang, Xiao and Chau, Michael},
 journal = {arXiv preprint arXiv:2405.14604},
 title = {A Watermark for Low-entropy and Unbiased Generation in Large Language Models},
 venue = {arXiv preprint arXiv:2405.14604},
 year = {2024}
}

@article{min2024imitate,
 author = {Min, Yingqian and Chen, Zhipeng and Jiang, Jinhao and Chen, Jie and Deng, Jia and Hu, Yiwen and Tang, Yiru and Wang, Jiapeng and Cheng, Xiaoxue and Song, Huatong and others},
 journal = {arXiv preprint arXiv:2412.09413},
 title = {Imitate, explore, and self-improve: A reproduction report on slow-thinking reasoning systems},
 venue = {arXiv preprint arXiv …},
 year = {2024}
}

@article{nie2024securing,
 author = {Nie, Hewang and Lu, Songfeng},
 journal = {Applied Intelligence},
 number = {21},
 pages = {10455--10472},
 publisher = {Springer},
 title = {Securing IP in edge AI: neural network watermarking for multimodal models},
 venue = {Applied Intelligence},
 volume = {54},
 year = {2024}
}

@article{peng2025skywork,
 author = {Peng, Yi and Wang, Xiaokun and Wei, Yichen and Pei, Jiangbo and Qiu, Weijie and Jian, Ai and Hao, Yunzhuo and Pan, Jiachun and Xie, Tianyidan and Ge, Li and others},
 journal = {arXiv preprint arXiv:2504.05599},
 title = {Skywork r1v: Pioneering multimodal reasoning with chain-of-thought},
 venue = {arXiv preprint arXiv …},
 year = {2025}
}

@inproceedings{qu2025provably,
 author = {Qu, Wenjie and Zheng, Wengrui and Tao, Tianyang and Yin, Dong and Jiang, Yanze and Tian, Zhihua and Zou, Wei and Jia, Jinyuan and Zhang, Jiaheng},
 booktitle = {34th USENIX Security Symposium (USENIX Security 25)},
 pages = {201--220},
 title = {Provably Robust Multi-bit Watermarking for $\{$AI-generated$\}$ Text},
 venue = {34th USENIX Security …},
 year = {2025}
}

@article{rastogi2024revisiting,
 author = {Rastogi, Saksham and Pruthi, Danish},
 journal = {arXiv preprint arXiv:2411.05277},
 title = {Revisiting the Robustness of Watermarking to Paraphrasing Attacks},
 venue = {arXiv preprint arXiv:2411.05277},
 year = {2024}
}

@article{raz2024authorship,
 author = {R{\"a}z, Tim},
 journal = {arXiv preprint arXiv:2403.06593},
 title = {Authorship and the Politics and Ethics of LLM Watermarks},
 venue = {arXiv preprint arXiv:2403.06593},
 year = {2024}
}

@article{tang2023watermarking,
 author = {Tang, Yuanmin and Yu, Jing and Gai, Keke and Qu, Xiangyan and Hu, Yue and Xiong, Gang and Wu, Qi},
 journal = {arXiv preprint arXiv:2311.05863},
 title = {Watermarking vision-language pre-trained models for multi-modal embedding as a service},
 venue = {arXiv preprint arXiv …},
 year = {2023}
}

@article{team2025kimi,
 author = {Team, Kimi and Du, Angang and Yin, Bohong and Xing, Bowei and Qu, Bowen and Wang, Bowen and Chen, Cheng and Zhang, Chenlin and Du, Chenzhuang and Wei, Chu and others},
 journal = {arXiv preprint arXiv:2504.07491},
 title = {Kimi-vl technical report},
 venue = {arXiv preprint arXiv …},
 year = {2025}
}

@article{tu2023waterbench,
 author = {Tu, Shangqing and Sun, Yuliang and Bai, Yushi and Yu, Jifan and Hou, Lei and Li, Juanzi},
 journal = {arXiv preprint arXiv:2311.07138},
 title = {Waterbench: Towards holistic evaluation of watermarks for large language models},
 venue = {arXiv preprint arXiv:2311.07138},
 year = {2023}
}

@article{wang2023towards,
 author = {Wang, Lean and Yang, Wenkai and Chen, Deli and Zhou, Hao and Lin, Yankai and Meng, Fandong and Zhou, Jie and Sun, Xu},
 journal = {arXiv preprint arXiv:2307.15992},
 title = {Towards codable watermarking for injecting multi-bits information to LLMs},
 venue = {arXiv preprint arXiv …},
 year = {2023}
}

@article{wang2024mitigating,
 author = {Wang, Xintong and Pan, Jingheng and Ding, Liang and Biemann, Chris},
 journal = {arXiv preprint arXiv:2403.18715},
 title = {Mitigating hallucinations in large vision-language models with instruction contrastive decoding},
 venue = {arXiv preprint arXiv:2403.18715},
 year = {2024}
}

@article{wang2025morphmark,
 author = {Wang, Zongqi and Gu, Tianle and Wu, Baoyuan and Yang, Yujiu},
 journal = {arXiv preprint arXiv:2505.11541},
 title = {Morphmark: Flexible adaptive watermarking for large language models},
 venue = {arXiv preprint arXiv:2505.11541},
 year = {2025}
}

@article{wei2025skywork,
 author = {Wei, Yichen and Peng, Yi and Wang, Xiaokun and Qiu, Weijie and Shen, Wei and Xie, Tianyidan and Pei, Jiangbo and Zhang, Jianhao and Hao, Yunzhuo and Song, Xuchen and others},
 journal = {arXiv preprint arXiv:2504.16656},
 title = {Skywork R1V2: Multimodal Hybrid Reinforcement Learning for Reasoning},
 venue = {arXiv preprint arXiv …},
 year = {2025}
}

@article{wu2023resilient,
 author = {Wu, Yihan and Hu, Zhengmian and Guo, Junfeng and Zhang, Hongyang and Huang, Heng},
 journal = {arXiv preprint arXiv:2310.07710},
 title = {A resilient and accessible distribution-preserving watermark for large language models},
 venue = {arXiv preprint arXiv:2310.07710},
 year = {2023}
}

@article{wu2024bypassing,
 author = {Wu, Qilong and Chandrasekaran, Varun},
 journal = {arXiv preprint arXiv:2403.14719},
 title = {Bypassing llm watermarks with color-aware substitutions},
 venue = {arXiv preprint arXiv:2403.14719},
 year = {2024}
}

@article{xie2024debiasing,
 author = {Xie, Yangxinyu and Li, Xiang and Mallick, Tanwi and Su, Weijie J and Zhang, Ruixun},
 journal = {arXiv preprint arXiv:2411.11203},
 title = {Debiasing watermarks for large language models via maximal coupling},
 venue = {arXiv preprint arXiv:2411.11203},
 year = {2024}
}

@article{xu2024freqmark,
 author = {Xu, Zhenyu and Zhang, Kun and Sheng, Victor S},
 journal = {arXiv preprint arXiv:2410.10876},
 title = {Freqmark: Frequency-based watermark for sentence-level detection of llm-generated text},
 venue = {arXiv preprint arXiv:2410.10876},
 year = {2024}
}

@article{xu2025majority,
 author = {Xu, Jiahao and Hu, Rui and Zhang, Zikai},
 journal = {arXiv preprint arXiv:2508.03829},
 title = {Majority Bit-Aware Watermarking For Large Language Models},
 venue = {arXiv preprint arXiv:2508.03829},
 year = {2025}
}

@article{yoo2023robust,
 author = {Yoo, KiYoon and Ahn, Wonhyuk and Jang, Jiho and Kwak, Nojun},
 journal = {arXiv preprint arXiv:2305.01904},
 title = {Robust multi-bit natural language watermarking through invariant features},
 venue = {arXiv preprint arXiv:2305.01904},
 year = {2023}
}

@article{yu2025saemark,
 author = {Yu, Zhuohao and Jiang, Xingru and Gu, Weizheng and Wang, Yidong and Zhang, Shikun and Ye, Wei},
 journal = {arXiv preprint arXiv:2508.08211},
 title = {SAEMark: Multi-bit LLM Watermarking with Inference-Time Scaling},
 venue = {arXiv preprint arXiv …},
 year = {2025}
}

@article{zhang2024personamark,
 author = {Zhang, Yuehan and Lv, Peizhuo and Liu, Yinpeng and Ma, Yongqiang and Lu, Wei and Wang, Xiaofeng and Liu, Xiaozhong and Liu, Jiawei},
 journal = {arXiv preprint arXiv:2409.09739},
 title = {PersonaMark: Personalized LLM watermarking for model protection and user attribution},
 year = {2024}
}

@inproceedings{zhang2024remark,
 author = {Zhang, Ruisi and Hussain, Shehzeen Samarah and Neekhara, Paarth and Koushanfar, Farinaz},
 booktitle = {33rd USENIX Security Symposium (USENIX Security 24)},
 pages = {1813--1830},
 title = {$\{$REMARK-LLM$\}$: A robust and efficient watermarking framework for generative large language models},
 venue = {33rd USENIX Security …},
 year = {2024}
}

@inproceedings{zhao2023protecting,
 author = {Zhao, Xuandong and Wang, Yu-Xiang and Li, Lei},
 booktitle = {International Conference on Machine Learning},
 organization = {PMLR},
 pages = {42187--42199},
 title = {Protecting language generation models via invisible watermarking},
 venue = {International Conference on …},
 year = {2023}
}

@misc{liu2021pretrainpromptpredictsystematic,
      title={Pre-train, Prompt, and Predict: A Systematic Survey of Prompting Methods in Natural Language Processing}, 
      author={Pengfei Liu and Weizhe Yuan and Jinlan Fu and Zhengbao Jiang and Hiroaki Hayashi and Graham Neubig},
      year={2021},
      eprint={2107.13586},
      archivePrefix={arXiv},
      primaryClass={cs.CL},
      url={https://arxiv.org/abs/2107.13586}, 
}

@misc{li2021prefixtuningoptimizingcontinuousprompts,
      title={Prefix-Tuning: Optimizing Continuous Prompts for Generation}, 
      author={Xiang Lisa Li and Percy Liang},
      year={2021},
      eprint={2101.00190},
      archivePrefix={arXiv},
      primaryClass={cs.CL},
      url={https://arxiv.org/abs/2101.00190}, 
}

@misc{tam2022parameterefficientprompttuningmakes,
      title={Parameter-Efficient Prompt Tuning Makes Generalized and Calibrated Neural Text Retrievers}, 
      author={Weng Lam Tam and Xiao Liu and Kaixuan Ji and Lilong Xue and Xingjian Zhang and Yuxiao Dong and Jiahua Liu and Maodi Hu and Jie Tang},
      year={2022},
      eprint={2207.07087},
      archivePrefix={arXiv},
      primaryClass={cs.CL},
      url={https://arxiv.org/abs/2207.07087}, 
}

@misc{yang2022robustprefixtuningtextclassification,
      title={On Robust Prefix-Tuning for Text Classification}, 
      author={Zonghan Yang and Yang Liu},
      year={2022},
      eprint={2203.10378},
      archivePrefix={arXiv},
      primaryClass={cs.CL},
      url={https://arxiv.org/abs/2203.10378}, 
}

@misc{lester2021powerscaleparameterefficientprompt,
      title={The Power of Scale for Parameter-Efficient Prompt Tuning}, 
      author={Brian Lester and Rami Al-Rfou and Noah Constant},
      year={2021},
      eprint={2104.08691},
      archivePrefix={arXiv},
      primaryClass={cs.CL},
      url={https://arxiv.org/abs/2104.08691}, 
}

@misc{liu2023gptunderstands,
      title={GPT Understands, Too}, 
      author={Xiao Liu and Yanan Zheng and Zhengxiao Du and Ming Ding and Yujie Qian and Zhilin Yang and Jie Tang},
      year={2023},
      eprint={2103.10385},
      archivePrefix={arXiv},
      primaryClass={cs.CL},
      url={https://arxiv.org/abs/2103.10385}, 
}

@misc{wang2025prefixtuningmodernizingprefixtuningdecoupling,
      title={Prefix-Tuning+: Modernizing Prefix-Tuning by Decoupling the Prefix from Attention}, 
      author={Haonan Wang and Brian Chen and Siquan Li and Xinhe Liang and Hwee Kuan Lee and Kenji Kawaguchi and Tianyang Hu},
      year={2025},
      eprint={2506.13674},
      archivePrefix={arXiv},
      primaryClass={cs.CL},
      url={https://arxiv.org/abs/2506.13674}, 
}

@inproceedings{ouyang-etal-2023-prefix,
    title = "On Prefix-tuning for Lightweight Out-of-distribution Detection",
    author = "Ouyang, Yawen  and
      Cao, Yongchang  and
      Gao, Yuan  and
      Wu, Zhen  and
      Zhang, Jianbing  and
      Dai, Xinyu",
    editor = "Rogers, Anna  and
      Boyd-Graber, Jordan  and
      Okazaki, Naoaki",
    booktitle = "Proceedings of the 61st Annual Meeting of the Association for Computational Linguistics (Volume 1: Long Papers)",
    month = jul,
    year = "2023",
    address = "Toronto, Canada",
    publisher = "Association for Computational Linguistics",
    url = "https://aclanthology.org/2023.acl-long.85/",
    doi = "10.18653/v1/2023.acl-long.85",
    pages = "1533--1545"
}

@misc{jia2022visualprompttuning,
      title={Visual Prompt Tuning}, 
      author={Menglin Jia and Luming Tang and Bor-Chun Chen and Claire Cardie and Serge Belongie and Bharath Hariharan and Ser-Nam Lim},
      year={2022},
      eprint={2203.12119},
      archivePrefix={arXiv},
      primaryClass={cs.CV},
      url={https://arxiv.org/abs/2203.12119}, 
}

@misc{liu2022fewshotparameterefficientfinetuningbetter,
      title={Few-Shot Parameter-Efficient Fine-Tuning is Better and Cheaper than In-Context Learning}, 
      author={Haokun Liu and Derek Tam and Mohammed Muqeeth and Jay Mohta and Tenghao Huang and Mohit Bansal and Colin Raffel},
      year={2022},
      eprint={2205.05638},
      archivePrefix={arXiv},
      primaryClass={cs.LG},
      url={https://arxiv.org/abs/2205.05638}, 
}

@INPROCEEDINGS{10657279,
  author={Tian, Xinyu and Zou, Shu and Yang, Zhaoyuan and Zhang, Jing},
  booktitle={2024 IEEE/CVF Conference on Computer Vision and Pattern Recognition (CVPR)}, 
  title={ArGue: Attribute-Guided Prompt Tuning for Vision-Language Models}, 
  year={2024},
  volume={},
  number={},
  pages={28578-28587},
  doi={10.1109/CVPR52733.2024.02700}}

@misc{yao2023visuallanguageprompttuningknowledgeguided,
      title={Visual-Language Prompt Tuning with Knowledge-guided Context Optimization}, 
      author={Hantao Yao and Rui Zhang and Changsheng Xu},
      year={2023},
      eprint={2303.13283},
      archivePrefix={arXiv},
      primaryClass={cs.CV},
      url={https://arxiv.org/abs/2303.13283}, 
}

@article{chen2025safeeraser,
 author = {Chen, Junkai and Deng, Zhijie and Zheng, Kening and Yan, Yibo and Liu, Shuliang and Wu, PeiJun and Jiang, Peijie and Liu, Jia and Hu, Xuming},
 journal = {arXiv preprint arXiv:2502.12520},
 title = {Safeeraser: Enhancing safety in multimodal large language models through multimodal machine unlearning},
 venue = {arXiv preprint arXiv …},
 year = {2025}
}

@article{hei2025unlocking,
 author = {Hei, Yonghua and Yan, Yibo and Liu, Shuliang and Zhou, Huiyu and Zhang, Linfeng and Hu, Xuming},
 journal = {arXiv preprint arXiv:2507.08603},
 title = {Unlocking Speech Instruction Data Potential with Query Rewriting},
 venue = {arXiv preprint arXiv …},
 year = {2025}
}

@article{hu2022hiure,
 author = {Hu, Xuming and Liu, Shuliang and Zhang, Chenwei and Li, Shuang and Wen, Lijie and Yu, Philip S},
 journal = {arXiv preprint arXiv:2205.02225},
 title = {Hiure: Hierarchical exemplar contrastive learning for unsupervised relation extraction},
 venue = {arXiv preprint arXiv …},
 year = {2022}
}

@article{huang2025video,
 author = {Huang, Yu and Chen, Junhao and Liu, Shuliang and Li, Hanqian and Zheng, Qi and Hu, Xuming and others},
 journal = {arXiv preprint arXiv:2506.00652},
 title = {Video Signature: In-generation Watermarking for Latent Video Diffusion Models},
 venue = {arXiv preprint arXiv …},
 year = {2025}
}

@article{huo2025pmark,
 author = {Huo, Jiahao and Liu, Shuliang and Wang, Bin and Zhang, Junyan and Yan, Yibo and Liu, Aiwei and Hu, Xuming and Zhou, Mingxun},
 journal = {arXiv preprint arXiv:2509.21057},
 title = {PMark: Towards Robust and Distortion-free Semantic-level Watermarking with Channel Constraints},
 venue = {arXiv preprint arXiv …},
 year = {2025}
}

@article{li2025treehop,
 author = {Li, Zhonghao and Zhang, Kunpeng and Ou, Jinghuai and Liu, Shuliang and Hu, Xuming},
 journal = {arXiv preprint arXiv:2504.20114},
 title = {TreeHop: Generate and Filter Next Query Embeddings Efficiently for Multi-hop Question Answering},
 venue = {arXiv preprint arXiv:2504.20114},
 year = {2025}
}

@article{ling2025wakenllm,
 author = {Ling, Zipeng and Tang, Yuehao and Liu, Shuliang and Yang, Junqi and Fu, Shenghong and Huang, Chen and Huang, Kejia and Wan, Yao and Hou, Zhichao and Hu, Xuming},
 journal = {arXiv preprint arXiv:2507.16199},
 title = {WAKENLLM: Evaluating Reasoning Potential and Stability in LLMs via Fine-Grained Benchmarking},
 venue = {arXiv preprint arXiv …},
 year = {2025}
}

@article{liu2025survey,
 author = {Liu, Shuliang and Liu, Hongyi and Liu, Aiwei and Duan, Bingchen and Zheng, Qi and Yan, Yibo and Geng, He and Jiang, Peijie and Liu, Jia and Hu, Xuming},
 journal = {arXiv preprint arXiv:2507.05288},
 title = {A Survey on Proactive Defense Strategies Against Misinformation in Large Language Models},
 venue = {arXiv preprint arXiv …},
 year = {2025}
}

@article{zhang2025bert,
 author = {Zhang, Junyan and Huang, Yiming and Liu, Shuliang and Gao, Yubo and Hu, Xuming},
 journal = {arXiv preprint arXiv:2505.18215},
 title = {Do BERT-Like Bidirectional Models Still Perform Better on Text Classification in the Era of LLMs?},
 venue = {arXiv preprint arXiv:2505.18215},
 year = {2025}
}

@article{zhang2025catmark,
 author = {Zhang, Yu and Liu, Shuliang and Yang, Xu and Hu, Xuming},
 journal = {arXiv preprint arXiv:2510.02342},
 title = {CATMark: A Context-Aware Thresholding Framework for Robust Cross-Task Watermarking in Large Language Models},
 venue = {arXiv preprint arXiv:2510.02342},
 year = {2025}
}

@article{zhang2025cohemark,
 author = {Zhang, Junyan and Liu, Shuliang and Liu, Aiwei and Gao, Yubo and Li, Jungang and Gu, Xiaojie and Hu, Xuming},
 journal = {arXiv preprint arXiv:2504.17309},
 title = {Cohemark: A novel sentence-level watermark for enhanced text quality},
 year = {2025}
}

@article{zhang2025unveiling,
 author = {Zhang, Junyan and Gao, Yubo and Yan, Yibo and Li, Jungang and Hou, Zhaorui and Tao, Sicheng and Liu, Shuliang and Dai, Song and Hei, Yonghua and Li, Junzhuo and others},
 journal = {arXiv preprint arXiv:2505.21191},
 title = {Unveiling Instruction-Specific Neurons \& Experts: An Analytical Framework for LLM's Instruction-Following Capabilities},
 venue = {arXiv preprint arXiv …},
 year = {2025}
}

@article{liu2026distilling,
  title={Distilling the Thought, Watermarking the Answer: A Principle Semantic Guided Watermark for Large Reasoning Models},
  author={Liu, Shuliang and Li, Xingyu and Liu, Hongyi and Yan, Yibo and Duan, Bingchen and Zheng, Qi and Fang, Dong and Su, Lingfeng and Hu, Xuming},
  journal={arXiv preprint arXiv:2601.05144},
  year={2026}
}

@article{liu2026vision,
  title={Vision-Language Introspection: Mitigating Overconfident Hallucinations in MLLMs via Interpretable Bi-Causal Steering},
  author={Liu, Shuliang and Yang, Songbo and Fang, Dong and Jia, Sihang and Tang, Yuqi and Su, Lingfeng and Peng, Ruoshui and Yan, Yibo and Zou, Xin and Hu, Xuming},
  journal={arXiv preprint arXiv:2601.05159},
  year={2026}
}
}


\clearpage
\appendix
\twocolumn[
    \begin{center}
        \textbf{\Large A Visual Semantic Adaptive Watermark grounded by Prefix-Tuning for Large Vision-Language Model} \\
        
        \vspace{0.5em} 
        
        {\large Supplementary Material}
        
        \vspace{1.0em} 
    \end{center}
]
In the supplementary materials, we report
\begin{itemize}
    \item Prefix-Tuning training setting and results (Appendix~\ref{app:prefix_tuning});
    \item Detailed ablation analysis (Appendix~\ref{app:ablation});
    \item Inference latency and algorithm efficiency analysis (Appendix~\ref{app:latency});
    \item Case Study (Appendix~\ref{app:case_study}).
\end{itemize}

\section{Prefix-Tuning Training Setting and Results}
\label{app:prefix_tuning}

\subsection{Training Configuration}
\label{sec:training_config}

\paragraph{Backbones and Data.} We train a dedicated prefix extractor for each backbone model (LLaVA-v1.5 and Qwen3-VL). We leverage the DCI dataset~\cite{urbanek2024pictureworth77text} as our external knowledge source, specifically utilizing its dense image-caption corpus to provide fine-grained visual supervision. Specifically, we randomly sampled a training set of 6,500 image-caption pairs to supervise the prefix optimization. For evaluation, we constructed a distinct, non-overlapping test set comprising 1,000 pairs.

\paragraph{Optimization Setup.} 
The prefix extractor is optimized using AdamW with a learning rate of $lr=2\times10^{-3}$, a batch size of 8, and a weight decay ($\ell_2$ regularization) of $10^{-4}$. The training is conducted for a total of 2,438 steps. Crucially, all parameters of the backbone LVLM remain frozen throughout this phase to ensure parameter efficiency.

\paragraph{Hyperparameter Settings.} 
We set the number of virtual prefix vectors to $L=84$. To provide a semantic prior, we employ a text-guided initialization strategy: the initial vectors are seeded with the embeddings of the prompt \enquote{The image shows}, while the remaining vectors are randomly initialized. Regarding the logit offset strength $\kappa$ in Eq.~\ref{eq:train_label}, we set $\kappa=10.0$ to align the magnitude of the learnable logit offsets with the original model logits.

\subsection{Training Dynamics and Efficiency}
\label{sec:training_results}

\begin{figure*}[h]
    \centering
    \includegraphics[width=1\linewidth]{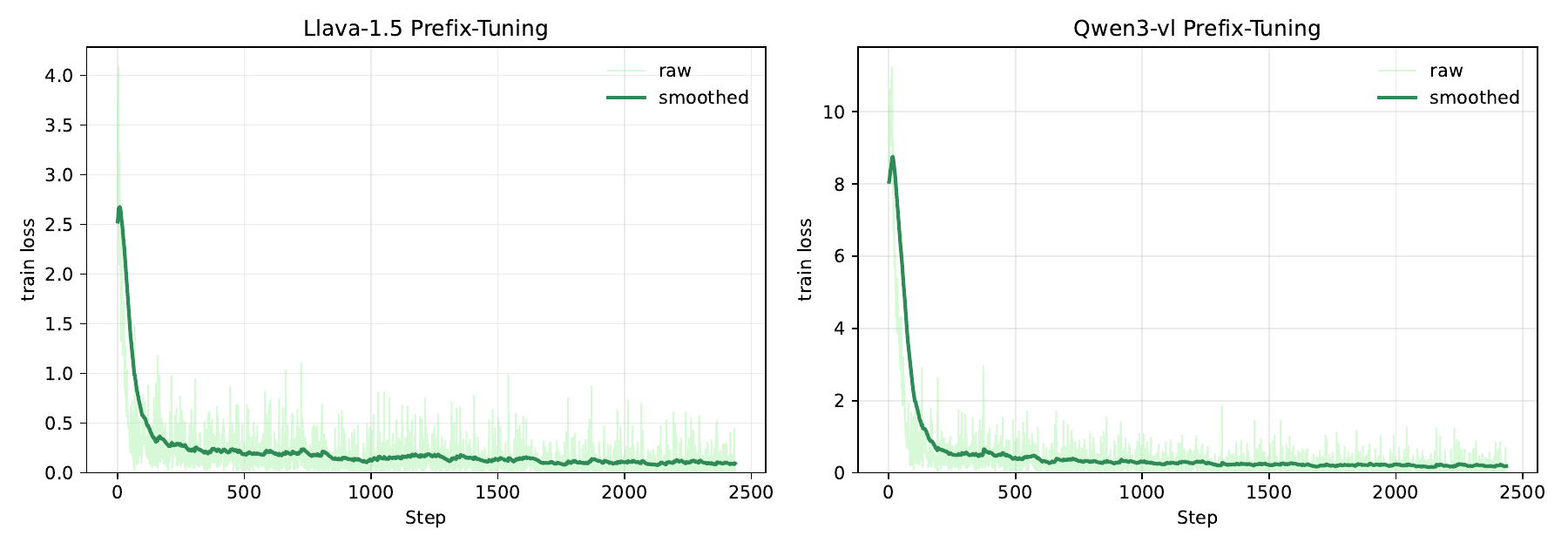}
    \caption{Training loss dynamics of the prefix-tuner on LLaVA-v1.5 (Left) and Qwen3-VL (Right) backbones over 2,438 steps (3 epochs). The light green lines represent the raw step-wise KL divergence loss, while the dark green lines depict the smoothed loss trajectory. Both models demonstrate rapid convergence in the early stages and maintain stability, validating the efficiency of our visual-evidence extraction learning.}
    \label{fig:loss}
    \vspace{-0.1in}
\end{figure*}

\paragraph{Computational Efficiency.} 
All experiments were conducted on a computational node equipped with 1 $\times$ NVIDIA A800-SXM4-80GB GPU. Despite the large scale of the backbone models, our lightweight prefix-tuning strategy demonstrates high training efficiency. The training phase for LLaVA-v1.5 was completed in approximately 7 hours, while the Qwen3-VL model required approximately 14 hours under identical hardware resources. This manageable overhead confirms the practicality of our extractor module.

\paragraph{Convergence Analysis.} 
To verify the effectiveness and stability of our training pipeline, we visualize the training loss curves for both backbones in Fig.~\ref{fig:loss}. As illustrated, both models exhibit a rapid convergence pattern: the KL divergence loss drops sharply within the initial training steps (approx. first 500 steps), indicating that the lightweight prefix-tuner quickly adapts to the visual-evidence extraction task. Following this rapid adaptation phase, the loss stabilizes at a low magnitude for the remainder of the 3 epochs. The raw loss fluctuations (light green) are typical for mini-batch optimization, while the smoothed curves (dark green) confirm a consistent downward trend, demonstrating that the prefix vectors have successfully learned to approximate the target dense visual distribution with high fidelity.

\subsection{Validation of Visual Evidence Weight Extraction}
\label{sec:vew_experiment}

\begin{table*}[t]
\centering
\caption{Effectiveness analysis of the Visual Evidence Weight Extraction module. We report the cosine similarity between extracted weights and ground-truth labels on the test set. The results show that our Prefix-Tuning strategy significantly outperforms both the raw vision-tower alignment and the static text prompting baseline, achieving high similarity after 3 epochs of training.}
\resizebox{0.85\linewidth}{!}{
    \begin{tabular}{lccccc}
    \toprule
     & \multicolumn{3}{c}{\textbf{Prefix-Tuning (Ours)}} & \multicolumn{2}{c}{Baselines}\\
    \cmidrule(lr){2-4}\cmidrule(lr){5-6}
     Model & Epoch 1 & Epoch 2 & \textbf{Epoch 3} & Vision-Tower Strategy & Prompting Strategy \\
    \midrule
    Llava-1.5 & 0.6467 & 0.7700 & \textbf{0.8022} & -0.6786 & 0.4744 \\
    Qwen3-VL  & 0.5608 & 0.5918 & \textbf{0.6143} & -0.4635 & 0.4133 \\
    \bottomrule
    \end{tabular}
}
\label{tab:vew_effectiveness}
\end{table*}

To strictly validate the efficacy of our training pipeline and the module's capability to extract meaningful Visual-Evidence Weights (VEW), we evaluated the performance evolution on the test set (1,000 samples) across training epochs. We employ \textbf{Cosine Similarity} as the primary metric to quantify the alignment between the extracted weights $\boldsymbol{\omega}$ and the ground-truth visual relevance distribution derived from dense captions.

\noindent\textbf{Baselines.} To establish a rigorous benchmark, we compare our trained prefix against two non-trained baselines:
\begin{itemize}
    \item \textbf{Vision-Tower Strategy:} This metric calculates the direct cosine similarity between the distinct visual embedding (from the pre-trained LVLM's vision encoder) and the vocabulary embeddings. This serves as a proxy for raw cross-modal alignment without LLM contextualization.
    \item \textbf{Prompting Strategy (Initialization):} This represents the zero-shot performance using only the initialization text ("The image shows") without the learned prefix vectors $\phi$. This isolates the gain achieved purely through prefix optimization.
\end{itemize}

\paragraph{Results Analysis.} 
As detailed in Table~\ref{tab:vew_effectiveness}, the results validate our training hypothesis. 
\textbf{(1) Training Progress:} Consistent with expectations, the similarity score improves steadily as training progresses. For LLaVA-1.5 and Qwen3-VL, the similarity peaks at \textbf{0.8022} and \textbf{0.6143} respectively at Epoch 3, demonstrating that the prefix successfully learns to map visual inputs to dense token-level evidence.
\textbf{(2) Comparison with Prompting:} The trained model at Epoch 3 significantly outperforms the Prompting Strategy (e.g., 0.8022 vs. 0.4744 on LLaVA). Even Epoch 1 surpasses the Prompting baseline, confirming that the learned soft prompts capture visual semantics far better than static text instructions.
\textbf{(3) Failure of Raw Vision Features:} The Vision-Tower Strategy yields negative values (e.g., -0.6786 on LLaVA). This indicates that raw cross-modal similarity contains significant noise and fails to represent the fine-grained, token-level evidence distribution required for watermarking. This underscores the necessity of our Prefix-Tuning approach, which leverages the LLM's internal knowledge to bridge the modality gap.

\section{Detailed Ablation Analysis}
\label{app:ablation}

In this section, we provide a granular analysis of the individual modules within VISA-Mark. While Sec.~4.2.2 focused on hyperparameter sensitivity ($\alpha$ and $\beta$), here we validate the architectural effectiveness of our framework: the strategy for visual evidence extraction and the structural necessity of our adaptive components.

\subsection{Ablation on Visual-Evidence Extraction Strategy}
\label{sec:ablation_vew_strategy}

\begin{table}[h]
\centering
\footnotesize 
\caption{\textbf{Ablation study on Visual-Evidence Extraction strategies.} We compare our learned Prefix-Tuning approach against raw feature alignment (Vision-Tower) and static text prompting. Our method achieves the optimal balance, delivering the lowest hallucination rate (Chair-I) and perplexity (PPL).}
\resizebox{1\linewidth}{!}{
    \begin{tabular}{lcccc}
    \toprule
    \thead{Ablation of\\VEW Extractor} & \thead{\textbf{Prefix-Tuning} \\ \textbf{(Ours)}} & \thead{Vision-Tower\\Strategy} & \thead{Prompting\\Strategy}  \\
    \midrule
     PPL $\downarrow$ & \textbf{5.52} & 5.75 &  5.61   \\
     BertScore & \textbf{93.07} & 92.48 &  92.51   \\
     Chair-I $\downarrow$ & \textbf{16.39} & 16.54 & 18.00   \\
    \bottomrule
    \end{tabular}
}
\label{tab:ablation_vew}
\end{table}

To validate the necessity of our learning-based \textbf{Visual Evidence Weight Extracting} module (Component $A$ in Sec.~3.2), we compared our \textbf{Prefix-Tuning} strategy against two alternative methods for acquiring Visual-Evidence Weights (VEW):
\begin{itemize}
    \item \textbf{Vision-Tower Strategy:} Directly computes the cosine similarity between the raw visual embedding (from the frozen vision encoder) and candidate token embeddings.
    \item \textbf{Prompting Strategy:} Utilizes the static text prompt \enquote{\textit{The image shows}} without trained prefix vectors to guide the probability distribution.
\end{itemize}

\paragraph{Analysis.} As presented in Table~\ref{tab:ablation_vew}, the \textbf{Prefix-Tuning} method yields superior performance across all metrics. 
\textbf{(1) Impact on Visual Fidelity:} Our method achieves the lowest hallucination rate (Chair-I: \textbf{16.39}), significantly outperforming the Prompting Strategy (18.00). This indicates that a simple text prompt fails to capture the fine-grained visual associations required to effectively guide the watermarking process against hallucinations.
\textbf{(2) Impact on Text Quality:} The Vision-Tower baseline results in the highest perplexity (PPL: 5.75). This suggests that raw visual embeddings, without the semantic adaptation provided by the LLM's prefix, contain cross-modal noise that disrupts the language model's fluency when used directly for logit perturbation.
\textbf{(3) Overall Superiority:} By bridging the modality gap through offline training, our Prefix-Tuning extractor successfully identifies high-quality visual evidence, enabling a watermarking mechanism that is both undetectable and visually faithful.

\subsection{Structural Ablation on Adaptive Components}
\label{sec:ablation_components}

\begin{table*}[t]
\centering
\footnotesize 
\caption{\textbf{Structural ablation of adaptive components.} We evaluate the impact of removing the entropy-aware mechanism (using fixed values) versus removing the component entirely. ``None'' denotes the full VISA-Mark framework. The results demonstrate that both the entropy-driven adaptation and the components themselves are crucial for minimizing perplexity and hallucinations (Chair-I).}
\resizebox{1\linewidth}{!}{
    \begin{tabular}{lccccc} 
    \toprule
    \multirow{2}{*}{\thead{Ablation of\\Components}} & \multirow{2}{*}{\textbf{None}} & \multicolumn{2}{c}{Uncertainty-based Vocabulary Partitioning} & \multicolumn{2}{c}{Evidence-Calibrated Logit Perturbation} \\
     \cmidrule(lr){3-4} \cmidrule(lr){5-6}
      & & w/o Entropy Mechanism & w/o Component & w/o Entropy Mechanism & w/o Component \\
    \midrule
    \midrule
     PPL $\downarrow$ & \textbf{5.52} & 5.62 &  5.69  & 5.81 & 5.58 \\
     BertScore & \textbf{93.07} & 92.32 &  92.51 & 92.21 & 92.45 \\
     Chair-I $\downarrow$ & \textbf{16.39} & 17.83 & 18.01 & 19.12 & 16.64 \\
    \bottomrule
    \end{tabular}
}
\label{tab:ablation_component}
\end{table*}

We further examine the structural contribution of the two core adaptive components: \textbf{Uncertainty-based Vocabulary Partitioning} (Component $B$ in Sec.~3.3) and \textbf{Evidence-Calibrated Logit Perturbation} (Component $C$ in Sec.~3.4). For each component, we performed two types of ablation:
\begin{itemize}
    \item \textbf{w/o Entropy Mechanism:} We deactivate the dynamic uncertainty regulation. Instead of adaptively scaling the partitioning ratio $\eta_t$ or the perturbation factor $\psi_t$ based on entropy, we apply fixed values derived from the average settings. This tests the hypothesis that watermarking strength should vary with model confidence.
    \item \textbf{w/o Component:} We completely remove the respective component from the pipeline to verify its holistic contribution.
\end{itemize}

\paragraph{Analysis.} The results in Table~\ref{tab:ablation_component} (where ``None'' represents the full VISA-Mark) reveal critical insights:
\textbf{(1) Necessity of Entropy Awareness:} Removing the entropy mechanism from either component leads to performance degradation. Notably, fixing the perturbation factor in Component $C$ causes a sharp increase in hallucinations (Chair-I rises from \textbf{16.39} to 19.12). This confirms that applying uniform/fixed perturbation without considering model uncertainty can force erroneous tokens in high-entropy states, whereas our adaptive mechanism successfully mitigates this risk.
\textbf{(2) Holistic Contribution:} Removing either component entirely (``w/o Component'') results in suboptimal text quality (higher PPL) and reduced visual consistency. The full VISA-Mark framework achieves the best synergy, validating that both vocabulary partitioning and logit perturbation are essential for the tripartite balance of text quality, visual fidelity, and detectability.

\section{Inference Latency and Algorithm Efficiency Analysis}
\label{app:latency}

\begin{table*}[t]
\centering
\caption{\textbf{End-to-end latency comparison.} Average generation time (seconds) for different watermarking methods generating 256 tokens. VISA-Mark maintains competitive efficiency compared to other semantic-aware methods (e.g., VLA).}
\resizebox{0.9\linewidth}{!}{
    \begin{tabular}{lccccccc}
    \toprule
    Model & \textbf{VISA-Mark} & VLA & KGW & SWEET & DiP & Unbiased & w/o watermark \\
    \midrule
    Llava-1.5 & 9.0387 & 9.4673 & 8.2615 & 8.2917 & 8.3464 & 8.3474 & 8.1646 \\
    Qwen3-VL  & 10.4423 & 11.3296 & 9.1579 & 9.1813 & 9.2829 & 9.1576 & 8.9892 \\
    \bottomrule
    \end{tabular}
}
\label{tab:latency}
\end{table*}

\begin{table*}[t]
\centering
\caption{\textbf{Component-wise latency breakdown.} Detailed overhead analysis (seconds) for VISA-Mark components under a 256-token setting. The \textit{Vocabulary Partitioning} represents the main computational cost, scaling with the model's vocabulary size ($|\mathcal{V}|$).}
\resizebox{1.0\linewidth}{!}{
    \begin{tabular}{lcccccc}
    \toprule
    \thead{Model} & \textbf{VISA-Mark} & \thead{Component A: \\ Visual Evidence\\Extracting} & \thead{Component B: \\ Uncertainty-based\\Vocabulary Partitioning} & \thead{Component C: \\ Evidence-Calibrated\\Logit Perturbation} & \thead{Total \\ Overhead} & \thead{Baseline \\ (w/o watermark)}\\
    \midrule    
    Llava-1.5 & 9.0387 & 0.2550 & 0.6830 & 0.0552 & 0.9985 & 8.1646 \\
    Qwen3-VL  & 10.4423 & 0.1455 & 1.1593 & 0.0589 & 1.3637 & 8.9892 \\
    \bottomrule
    \end{tabular}
}
\label{tab:latency-breakdown}
\end{table*}

Table~\ref{tab:latency} quantifies the end-to-end generation latency across two LVLMs under standardized conditions (256 generated tokens). While VISA-Mark introduces a moderate latency increase compared to lightweight baselines like KGW, the additional overhead is manageable (e.g., approx. +0.87s on LLaVA-1.5 and +1.45s on Qwen3-VL relative to the unwatermarked baseline). This trade-off is justified by the significant gains in vision-aligned semantic consistency.

To pinpoint computational bottlenecks, we provide a granular component-wise breakdown in Table~\ref{tab:latency-breakdown}. 
Notably, the \textit{Visual Evidence Extracting} incurs negligible overhead (0.26s for LLaVA, 0.15s for Qwen). Since this prefix-based extraction is computed only once per image input, its cost is amortized across the entire generation process, remaining invariant to the output sequence length.

\paragraph{Bottleneck Analysis.}
The primary source of latency is the \textit{Uncertainty-based Vocabulary Partitioning} component (0.68s for LLaVA vs. 1.16s for Qwen). This disparity is directly attributable to the algorithmic complexity of the dynamic partitioning mechanism. Unlike static hashing in KGW ($O(1)$), our method necessitates calculating and sorting visual relevance scores across the candidate vocabulary $\mathcal{V}$ at each step. The time complexity of this operation is approximately $O(|\mathcal{V}|\log(|\mathcal{V}|))$.
Consequently, Qwen3-VL, which operates on a significantly larger vocabulary ($\sim$152k tokens) compared to LLaVA-1.5 ($\sim$32k tokens), exhibits a proportionally higher latency in this component. Despite this, the overall efficiency remains within a practical range for offline generation tasks.

\section{Case Study}
\label{app:case_study}

\begin{figure*}[h]
    \centering
    \includegraphics[width=1\linewidth]{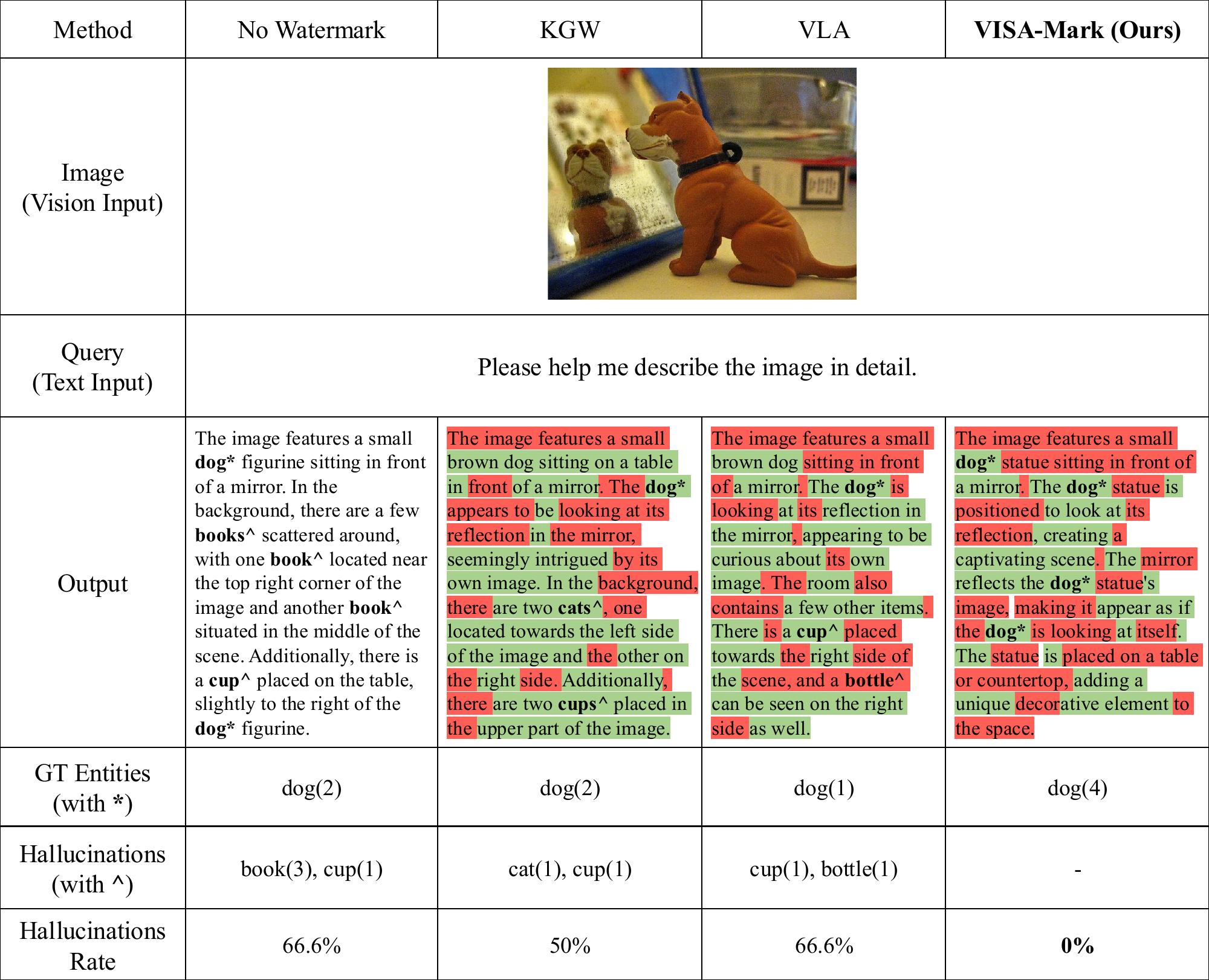}
    \caption{Qualitative comparison of watermarked responses on sample \texttt{COCO\_val2014\_000000475928}. Green and red highlights denote watermarked (green-list) and unwatermarked (red-list) tokens, respectively. \textbf{Bold} terms represent the detected object entities, where `*' marks ground-truth visual evidence and `\^{}' marks hallucinations. While baseline methods (KGW, VLA) and even the unwatermarked model produce hallucinations (e.g., non-existent ``cats'' or ``cups''), \textbf{VISA-Mark} successfully generates a hallucination-free description (0\% rate) with all correct entities watermarked. This demonstrates our framework's ability to align watermark injection with visual evidence, effectively correcting model-intrinsic errors.}
    \label{fig:case_study}
    \vspace{-0.1in}
\end{figure*}

To intuitively demonstrate the efficacy of VISA-Mark in preserving visual fidelity, we present a detailed qualitative comparison in Fig.~\ref{fig:case_study} (Sample ID: \texttt{COCO\_val2014\_000000475928}). The figure visualizes the generated descriptions from the unwatermarked baseline, KGW~\cite{kirchenbauer2023watermark}, VLA-Mark~\cite{liu2025vlamarkcrossmodalwatermark}, and our VISA-Mark. Green and red highlights indicate whether a token was successfully embedded with the watermark signal (i.e., selected from the green list).

\paragraph{Baseline Failures.} 
As observed, standard methods struggle to maintain visual grounding.
\begin{itemize}
    \item \textbf{Vision-Agnostic Failure (KGW):} The KGW method introduces a severe hallucination—a ``cat'' appearing in the reflection. This likely occurs because the correct token (``dog'') was randomly assigned to the red list. The rigid, vision-agnostic partitioning suppressed the correct visual evidence, forcing the model to select a semantically related but visually incorrect alternative (``cat'') that happened to be in the green list.
    \item \textbf{Visual Noise Interference (VLA):} While VLA attempts to incorporate visual features, it hallucinates a ``cup'' and a ``bottle.'' This suggests that directly injecting global visual features without filtering can introduce background noise or misalignments, causing the model to misinterpret ambiguous regions.
    \item \textbf{Intrinsic Model Hallucinations:} Notably, even the unwatermarked baseline hallucinates ``books'' and a ``cup.'' This indicates that the base LVLM has inherent uncertainty in this complex scene (a dog looking into a mirror). Standard watermarks fail to correct—and often exacerbate—these intrinsic errors.
\end{itemize}

\paragraph{VISA-Mark Superiority.}
In stark contrast, \textbf{VISA-Mark} generates a completely accurate description with a \textbf{0\% hallucination rate}. It correctly identifies the ``dog'' without fabricating non-existent objects. This success stems from the discriminative power of our \textbf{Visual-Evidence Weight (VEW) Extractor}, which functions as both a promoter of truth and a suppressor of error.
By explicitly quantifying evidentiary support, our mechanism grants high weights to visually grounded tokens (``dog''), ensuring their inclusion in the green list via \textit{Uncertainty-based Partitioning} and enhancing their likelihood via \textit{Calibrated Perturbation}.
Simultaneously, it implicitly penalizes hallucinated tokens (e.g., ``cat'', ``cup'') by assigning them low visual relevance scores. Unlike vision-agnostic methods that might randomly boost these errors, VISA-Mark denies them the adaptive logit enhancement, thereby significantly reducing their sampling probability. This bidirectional guidance effectively anchors the model to the visual reality, mitigating both intrinsic model uncertainty and watermark-induced noise.

\end{document}